\documentclass[aoas]{imsart}

\RequirePackage{amsthm,amsmath,amsfonts,amssymb}
\RequirePackage[authoryear]{natbib}
\RequirePackage{graphicx}
\RequirePackage{bm}
\RequirePackage{url}
\RequirePackage{xcolor}
\RequirePackage[normalem]{ulem}
\newcommand{\fclnladd}[1]{{\color{black}#1}}
\newcommand{\added}[1]{\textcolor{black}{#1}}
\newcommand{\revblue}[1]{\textcolor{black}{#1}}
\newcommand{\revgreen}[1]{\textcolor{black}{#1}}

\newcommand{\theoremexplain}[1]{#1}
\definecolor{FCLRevisionGreen}{RGB}{0,0,0}
\startlocaldefs
\newcommand{\indep}{\mathrel{\perp\!\!\!\perp}}
\theoremstyle{plain}
\newtheorem{theorem}{Theorem}[section]
\newtheorem{proposition}[theorem]{Proposition}
\newtheorem{corollary}[theorem]{Corollary}

\theoremstyle{definition}
\newtheorem{definition}[theorem]{Definition}
\newtheorem{assumption}[theorem]{Assumption}
\theoremstyle{definition}

\endlocaldefs

\begin{document}

\begin{frontmatter}
\title{Fisher Contrastive Learning to Mitigate Feature Suppression: A Case Study in Acute Promyelocytic Leukemia Diagnosis}
\runtitle{Fisher Contrastive Learning for APL Diagnosis}

\begin{aug}
\author[A,B]{\fnms{Shushan}~\snm{Wu}}
\author[C]{\fnms{Yongkai}~\snm{Chen}}
\author[A]{\fnms{Xilin}~\snm{Gong}}
\author[A]{\fnms{Yufang}~\snm{Liu}}
\author[A]{\fnms{Wenxuan}~\snm{Zhong}\ead[label=e1]{wenxuan@uga.edu}}
\author[A]{\fnms{Ping}~\snm{Ma}\ead[label=e2]{pingma@uga.edu}}

\address[A]{Department of Statistics, University of Georgia%
  \printead[presep={,\ }]{e1}%
  \printead[presep={;\ }]{e2}}
\address[B]{Department of Radiology, University of North Carolina at Chapel Hill}
\address[C]{Department of Statistics, Harvard University}
\end{aug}

\begin{abstract}
Self-supervised contrastive learning (SSCL) learns representations by contrasting augmented views of the same sample, but it can exhibit feature suppression, where dominant or easy-to-learn factors suppress other class-relevant signals, degrading the performance of representations, particularly in high-dimensional data with heterogeneous features. We introduce Fisher Contrastive Learning (FCL), which uses a regularized Fisher trace criterion to address feature suppression. FCL learns embeddings by maximizing the discriminative strength of Fisher discriminant functions, producing representations that preserve class-relevant information implied by the augmentation process. We prove that the representations learned by FCL could alleviate the feature suppression effects. We demonstrate FCL on simulated benchmarks designed to induce feature suppression and on a real-world biomedical application, pretraining on large white blood cell microscopy image dataset and transfer learning for diagnosis of acute promyelocytic leukemia in an external cohort. In all experiments, FCL outperforms the self-supervised baselines and yields more robust and discriminative representations. 
\end{abstract}

\end{frontmatter}

\section{Introduction}
\label{sec:intro}
% sscl growing importance in applications like image recognition, natural language processing, and pretraining
\revgreen{Accurate diagnosis of hematologic malignancies from peripheral blood smear images is a critical yet challenging task in clinical practice. In particular, acute promyelocytic leukemia (APL) is a rare but life-threatening subtype of acute myeloid leukemia (AML) that requires rapid identification, since delayed treatment can lead to fatal hemorrhagic complications. Microscopy-based APL diagnosis is challenging because APL shares substantial morphological similarity with other leukemias and annotated APL images are scarce.}
These challenges motivate the development of robust representation learning methods that can leverage large collections of unlabeled white blood cell (WBC) images to improve diagnostic performance on limited labeled APL datasets.

In our study, we use two WBC image resources. \revblue{The first is a large, public WBC microscopy image dataset \citep{tsutsui2023wbcatt}.} The second is a much smaller, clinically focused APL dataset \citep{sidhom2021deep}, in which each cell inherits the molecular diagnosis of its source patient (APL or other AML).
\revblue{Our goal is to leverage the large WBC dataset for pretraining representations that may be better suited to diagnostic learning with limited labeled patients, and fine-tune the pretrained model on the small APL dataset.} This motivates a self-supervised contrastive pretraining stage on WBC images, followed by supervised fine-tuning for APL diagnosis.

Self-supervised learning is a powerful machine learning method in which a model learns from unlabeled data by creating its own labels from the inherent structure of the data.
Self-supervised representation learning includes contrastive and non-contrastive approaches. Self-supervised contrastive learning (SSCL) \citep{chen2020simple, robinson2020contrastive, kalantidis2020hard} brings augmented views of the same sample closer in an embedding space while pushing apart views from different samples.
Other approaches learn from augmented views without explicit negative-pair repulsion, including clustering-based methods \citep{caron2020unsupervised} and SimSiam \citep{chen2021exploring}. By exploiting abundant unlabeled data, self-supervised learning reduces the dependence on large annotated datasets and helps alleviate the annotation bottleneck.
%SSCL offers a paradigm shift from traditional supervised learning by leveraging unlabeled data to generate useful representations. This approach addresses the annotation bottleneck, a significant barrier in deploying deep learning models, by %eliminating the need 
%reducing the reliance on large annotated datasets. %SSCL has great potential in

% SSCL demonstrates immense potential in various applications, including 
% multi-modal data integration \citep{huang2023visual,  chen2024stimage}, exemplified by models such as Contrastive Language-Image Pre-Training (CLIP) model \citep{radford2021learning}. It facilitates learning domain-invariant features from data with different distributions \citep{hjelmlearning, kim2021selfreg, wei2023domain}, and serves as a pre-training method, reducing the need for labeled data for downstream tasks \citep{chen2020simple, gao2021simcse}, served as a dimension reduction method in biomedical case-control studies \citep{abid2018exploring, hawke2024contrastive}. In addition, self-supervised contrastive learning can function as a clustering method in spatial transcriptomics, where the derived cell representations reveal global patterns \citep{liu2024spagic, ren2022identifying}. %and the cell representations show global patterns.}

Contrastive learning has been used across a wide range of settings. It supports multi-modal representation learning and data integration \citep{huang2023visual, chen2024stimage}, exemplified by Contrastive Language–Image Pre-Training (CLIP) \citep{radford2021learning}. It can also learn features that transfer across heterogeneous or shifted distributions, enabling more domain-invariant representations \citep{hjelmlearning, kim2021selfreg, wei2023domain}. In natural language processing and computer vision, SSCL is widely used for pre-training to improve downstream performance with limited labels \citep{chen2020simple, gao2021simcse}. In biomedical applications, related contrastive objectives have been used for dimension reduction in case--control studies \citep{abid2018exploring, li2024probabilistic, zhang2025contrastive, hawke2024contrastive}; this case--control formulation differs from the augmented-instance formulation considered here. Contrastive learning has also been used for unsupervised discovery tasks such as clustering in spatial transcriptomics, where learned cell embeddings can reveal global structure \citep{guojade, liu2024spagic, ren2022identifying}.

\subsection{\revgreen{Feature Suppression}}
\revblue{Standard SSCL objectives can exhibit feature suppression, in which dominant or easy-to-learn factors limit the recovery of other signals useful for downstream prediction \citep{chen2021intriguing}.} In conventional SSCL, augmented views are constructed to preserve ``content'' while perturbing ``style,'' and the learning objective encourages embeddings of augmented views from the same cell to be close while pushing apart embeddings from different cells.

This feature suppression phenomenon has been documented and analyzed in recent work on contrastive learning. Dominant features (e.g., color or global texture) can overshadow other semantically meaningful factors (e.g., object class or shape) in the learned embedding \citep{rusak2022content}. Easy-to-learn shared features between augmented pairs can suppress more complex or fragile features \citep{chen2021intriguing}, and imperfect augmentations that preserve class-irrelevant structure further exacerbate the problem \citep{xue2023features}. Dimensional collapse is a related but distinct problem: learned embeddings occupy a lower-dimensional subspace of the available embedding space \citep{jing2022understanding}. Feature suppression can also affect multi-attribute tasks, where learning one attribute may obscure information relevant to another \citep{li2023addressing}. \revblue{In the WBC transfer-learning setting, the relevant question is whether the pretrained representation supports diagnostic prediction from a small labeled patient cohort.}

%dimension collapse. 
%The resulting lack of discrimination power fails to capture the full complexity and richness of the data, degrading the performance of downstream tasks. Severe feature suppression effects in SSCL require effective strategies with theoretical guarantees to mitigate this critical problem.

% identify the limitation of the current solutions, lack of statistical rigor and theoretical guarantees; they only offered heuristic-based solutions or empirical approaches in SSCL

% Previous research has not defined feature suppression effect mathematically. Now we formulate the definition of feature suppression effects. 

\revgreen{These concerns motivate the study of pretrained representation geometry for downstream APL classification. With few labeled patients, diagnostic performance may depend on how readily a supervised classifier can use the learned representation. We hypothesize that organizing between-instance variation relative to within-instance augmentation variability makes representations better suited to diagnostic learning in this limited-sample setting.}

\begin{figure}[htbp]
\centering
\includegraphics[width=0.8\textwidth]{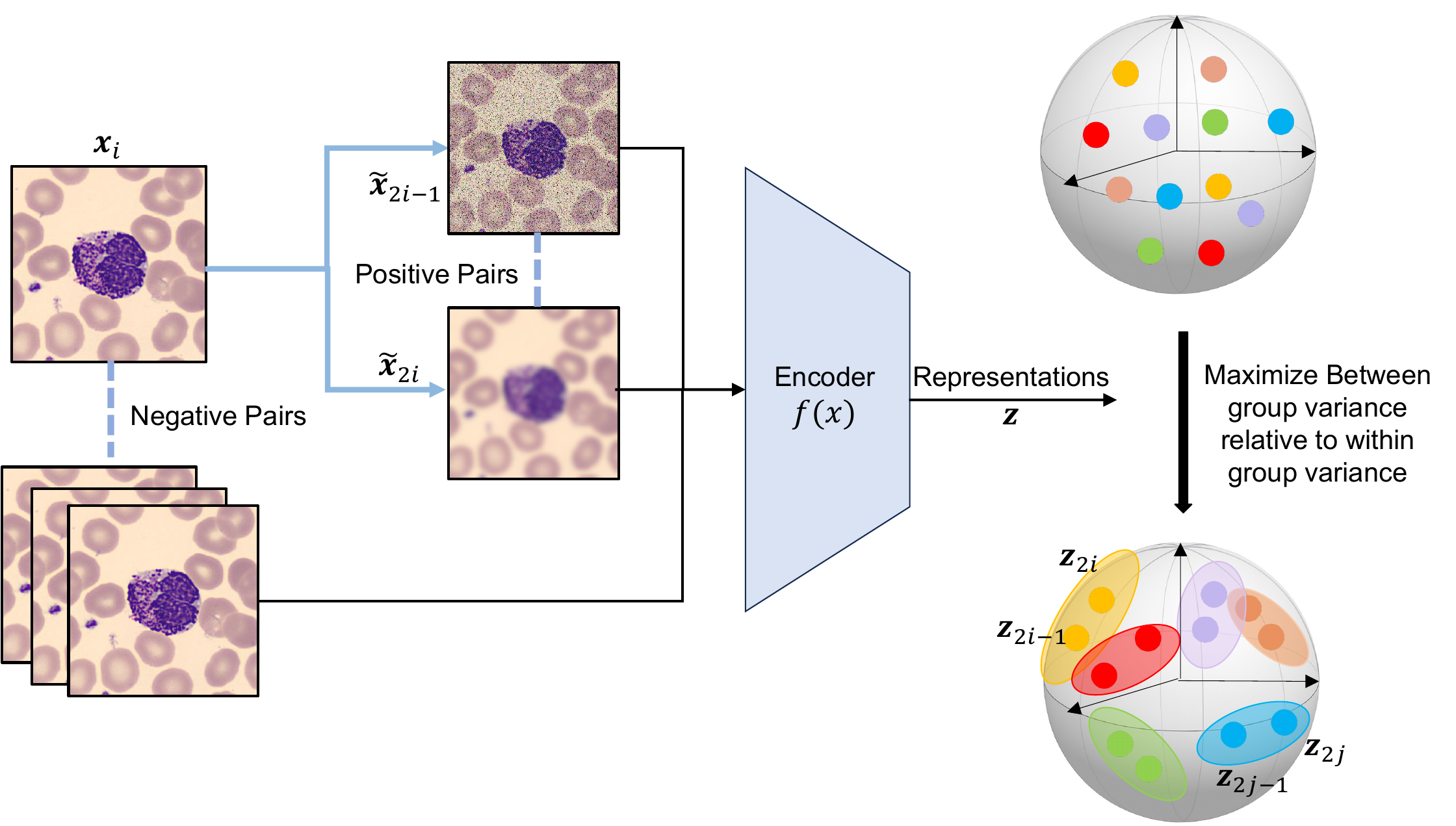}

\caption{Overview of the Fisher Contrastive Learning framework.
An input image is augmented into two views that form a positive pair, while other images in the batch serve as separate negative classes. An encoder projects all views into an embedding space, where FCL reframes the task as an
N-class problem and minimize the variance of positive pairs while maximizing variance between different image classes. }\label{fig:model_fig}
\end{figure}

\revblue{Approaches addressing feature suppression or the related problem of dimensional collapse include implicit feature modification \citep{robinson2021can}, changes to embedding dimension or augmentation noise \citep{xue2023features}, and DirectCLR, which applies InfoNCE to a fixed subvector of the representation without a trainable projector \citep{jing2022understanding}. Existing theory provides feature-retention guarantees under specific data, model, and optimization assumptions. For example, \citet{xue2023features} analyze minimum-norm solutions and a joint supervised--unsupervised contrastive loss. Here we study how the regularized Fisher trace criterion controls linear recovery uniformly over normalized linear combinations in a selected signal space, under the normalization and spectral assumptions stated below.}

\subsection{\revgreen{Fisher Contrastive Learning}}
% we borrow a tool of SDR and solve the problem
In this work, we address feature suppression in SSCL by reformulating contrastive learning as a nonlinear dimension reduction problem \citep{li1991sliced,cook1994using}. For a minibatch of $N$ original images, two augmented views of each image yield $2N$ samples and define an $N$-class classification problem, with each original image corresponding to one augmented instance class. FCL then optimizes a Fisher contrastive loss in the embedding space. Specifically, FCL constructs within-class and between-class scatter matrices from the augmented-view embeddings and maximizes a regularized Fisher trace criterion, thereby favoring small within-instance variation and large between-instance separation, as shown in Figure~\ref{fig:model_fig}. %Viewed through the lens of sufficient dimension reduction \citep{lee2013general, li2018sufficient},
\revblue{Our analysis identifies the population target of this minibatch criterion, quantifies finite-batch approximation, and distinguishes linear feature accessibility from sufficiency at the level of $\sigma$-fields. A central-projection theorem connects the SDR (sufficient dimension reduction) formulation to the original FCL score under an explicit between-source signal condition.}

\revgreen{Our contributions are threefold. First, we introduce a two-view Fisher trace criterion for SSCL, identify its fixed-encoder population limit, and derive finite-batch approximation bounds. We establish a central-projection result under a between-source signal condition and uniform linear-recovery bounds over a selected signal eigenspace under population covariance normalization and spectral separation. Second, a controlled simulation compares class separation and downstream prediction from the FCL and SimCLR representations. Third, we evaluate transfer from self-supervised WBC pretraining to APL diagnosis using patient-level outcomes in a held-out prospective cohort.}
%(1) we introduce FCL, a nonlinear sufficient dimension reduction method for SSCL, that offers a robust solution to the feature suppression effects by learning an exhaustive and unbiased function class; (2) FCL prevents the low embedding dimensionality by maximizing the discrimination power of the Fisher discriminant functions; (3) we demonstrate the effectiveness of the proposed method on various datasets that exhibit feature suppression effects and benchmark image datasets compared with other self-supervised learning methods.

We organize the paper as follows. In Section~\ref{sec:supp}, we introduce the white blood cell image datasets for pretraining and a small cell-image dataset from patients with APL for classification.
In Section~\ref{sec:pre}, we provide preliminaries of self-supervised learning and formally define feature suppression. In Section~\ref{sec:fcl}, we introduce Fisher Contrastive Learning. %In Section \ref{sec:thm}, we outline the theoretical foundation in sufficient dimension reduction and its theoretical contribution to SSCL. 
In Section~\ref{sec:sim}, we evaluate the method in a controlled simulation.
We then present the white blood cell application in Section~\ref{sec:exp}. This is followed by conclusions in Section~\ref{sec:con}, where we discuss the results, limitations, and potential directions for methodological development and applications. \revblue{The supplement provides proofs of the population and finite-batch results, the SDR analysis, and comparisons with related objectives.}

\section{\revblue{White Blood Cell Microscopy Image Datasets for Diagnostic Transfer}}\label{sec:supp}
% \paragraph{Motivating Example.}  %We further checked the rank of the learned embeddings and the rank is consistent with the true rank of the class relevant features. 

\revblue{We investigate self-supervised contrastive learning (SSCL) for diagnostic transfer to acute promyelocytic leukemia (APL) from peripheral blood smears.} Our analysis leverages two complementary white blood cell (WBC) image datasets. \revblue{The first is a large public WBC dataset that we treat as an unlabeled corpus for SSCL pretraining.} The second is a small, clinically curated APL dataset, which serves as the primary application dataset for assessing diagnostic performance. \revblue{In this section, we describe the imaging data that motivate diagnostic learning with limited labeled patients.}

\begin{figure}[htbp]
\centering
\includegraphics[width=0.9\textwidth]{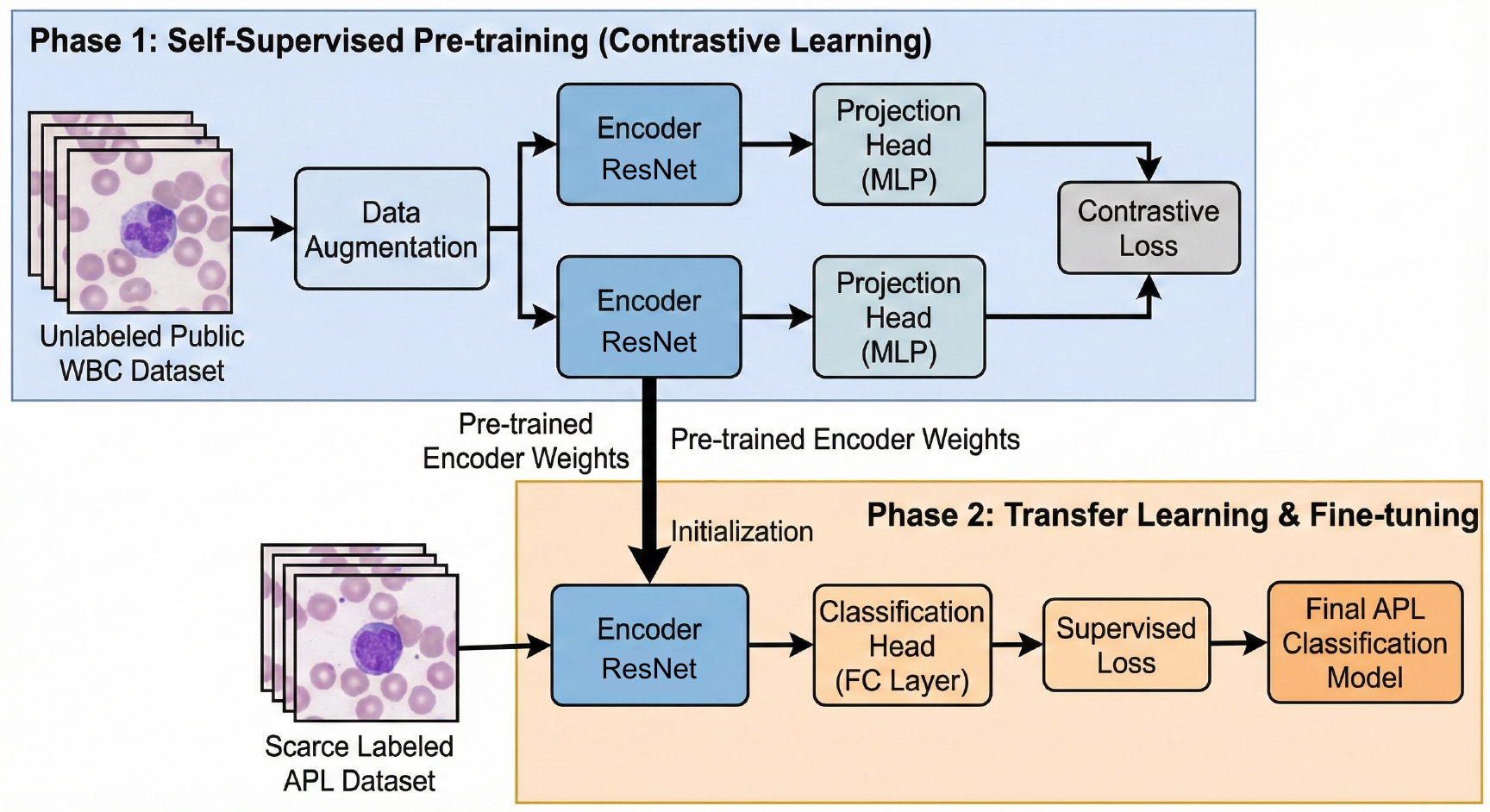}

\caption{Workflow of the proposed two-phase transfer-learning framework for APL diagnosis. \revblue{Phase~1 uses self-supervised contrastive learning on a large public WBC dataset to pretrain a ResNet encoder.} Phase~2 initializes the encoder with the pretrained weights and fine-tunes it on the smaller labeled APL dataset using a supervised loss.}\label{fig:workflow}
\end{figure}

\subsection{Large White Blood Cell Dataset for Pretraining}\label{subsec:wbc_data}
\revblue{The pretraining dataset is a comprehensive WBC microscopy image collection from \citet{tsutsui2023wbcatt}.} It contains 10{,}298 single-cell images, with 6{,}169, 1{,}030, and 3{,}099 images in the published training, validation, and test splits, respectively. The five cell types are basophils, eosinophils, lymphocytes, monocytes, and neutrophils.

\revblue{In our workflow (Figure~\ref{fig:workflow}), we use this collection as a pretraining dataset.} We apply standard SSCL augmentations such as random cropping, color jittering, and noise, and use a ResNet encoder to map each augmented view into a low-dimensional embedding space.

\revblue{In our application, the WBC dataset provides a large unlabeled corpus for SSCL pretraining, enabling us to learn transferable WBC representations without relying on exhaustive expert labels. We assess their downstream utility through patient-level APL classification, as summarized in Section~\ref{sec:exp}.}

% \url{https://paperswithcode.com/dataset/cub-200-2011}

% To address this, we propose a novel statistical framework that reformulates SSCL as a dimension reduction problem. FCL leverages Fisher discriminant functions to preserve the sufficient dimension reduction functional class, mitigating feature suppression effects with theoretical guarantees and enhancing representation learning. Our method demonstrates effectiveness across both simulated and real-world datasets (detailed in Sections \ref{sec:sim} and \ref{sec:exp}). By integrating Fisher discriminant analysis into SSCL, FCL provides a robust solution to the challenges of representation learning, bridging the gap between contrastive learning and statistical theory to advance the field (detailed in the Section \ref{sec:fcl}).

\subsection{Small APL Dataset for Classification}\label{subsec:apl_data}

Acute promyelocytic leukemia (APL) is a rare but life-threatening subtype of acute myeloid leukemia and is considered an oncologic emergency. If untreated, APL can lead to rapid, fatal hemorrhage, whereas timely recognition enables highly effective targeted therapy with one of the highest cure rates among leukemias \citep{sidhom2021deep}. \revblue{In many settings, particularly where access to molecular testing (e.g., PCR-based assays) is limited or delayed, clinicians rely heavily on peripheral-smear assessment to identify suspected APL cases.}

The APL dataset of \citet{sidhom2021deep} consists of peripheral-smear images annotated at the cell and patient levels. For each patient, multiple WBCs are imaged, and each patient is labeled as APL or other AML based on expert review and confirmatory molecular testing. The published dataset contains 106 patients: a discovery cohort of 82 patients (22 APL and 60 other AML) for model development and a prospective cohort of 24 patients (12 APL and 12 other AML) for held-out evaluation. The small number of patients, particularly in the APL class, creates high sampling variability and motivates transfer learning.

The challenge of lacking training samples motivates the use of SSCL as a pre-training strategy. \revblue{Our goal is to leverage the large WBC dataset described in Section~\ref{subsec:wbc_data} to learn representations that may be better suited to diagnostic learning with limited labeled patients.} We then transfer these representations to the APL dataset via a two-phase framework (Figure~\ref{fig:workflow}). In Phase~1, we pretrain the encoder on the unlabeled WBC images using SSCL, comparing a standard contrastive baseline (SimCLR) with the proposed Fisher Contrastive Learning. In Phase~2, we initialize the encoder  with the pretrained weights, attach a supervised classification head, and fine-tune the model on the discovery cohort to predict APL versus non-APL status at patient level.

\revblue{The APL dataset provides a clinical testbed for evaluating the usefulness of pretrained representations when diagnostic labels are available for relatively few patients. The prospective-cohort evaluation assesses downstream prediction.}

\section{Preliminaries}\label{sec:pre}
\subsection{SimCLR}
SimCLR \citep{chen2020simple} is a widely used SSCL method and serves as the principal contrastive baseline in this paper.
Suppose that the dataset contains $N$ data points. 
%For each feature vector $\boldsymbol{x}_k$ and its classification label $y_k$, we assume $\boldsymbol{x}=\boldsymbol{u}+\boldsymbol{\epsilon}$, where $\boldsymbol{u}=(y\phi_0+\mu_0)\boldsymbol{v}_0+(\rho_1\phi_1+\mu_1)\boldsymbol{v}_1+\cdots+(\rho_r\phi_r+\mu_r)\boldsymbol{v}_r$, $\boldsymbol{v}_0$ is class-relevant feature vector, and $\left\{\boldsymbol{v}_{1},\cdots, \boldsymbol{v}_r\right\}$ are class-irrevlevant feature vectors, $\boldsymbol{\epsilon}$ and $\boldsymbol{\rho}_{\cdot}$ are Gaussian random noise, $\left\{\mu_0,...,\mu_r\right\}$ are group means, and $\left\{\phi_0,...,\phi_r\right\}$ are constants that indicate the strength of each feature. Given a feature vector, $\boldsymbol{x}=\boldsymbol{u}+\boldsymbol{\epsilon}$, we apply the data augmentation so that the augmented feature is $\tilde{\boldsymbol{x}}=\boldsymbol{u}+\tilde{\boldsymbol{\epsilon}}$. In our motivating example, $X_1$ and $X_2$ consists of the class-relevant characteristic $\boldsymbol{v}_0$, and $\left\{X_3,...,X_{18}\right\}$ constitutes the class-irrelevant features $\left\{\boldsymbol{v}_1,...,\boldsymbol{v}_r\right\}$. 
For each input $\boldsymbol{x}_k$, two transformations are sampled independently from the augmentation distribution conditional on that source image, yielding views $\tilde{\boldsymbol{x}}_{2k-1}$ and $\tilde{\boldsymbol{x}}_{2k}$. These $2N$ views form $N$ positive pairs; views derived from different source images are treated as negatives. We use $Y$ to denote the augmented instance label that identifies the source image within the batch.
The nonlinear encoder $f:\Omega_X\rightarrow\mathbb{R}^d$ maps an augmented input $\tilde{\boldsymbol{x}}_i\in\mathbb{R}^p$ to an embedding $\boldsymbol{z}_i=f(\tilde{\boldsymbol{x}}_i)\in\mathbb{R}^d$. Here $\Omega_X$ denotes the augmented-input space, also written $\mathcal X$ below; $p$ and $d$ are the input and embedding dimensions.
%In supervised contrastive leaning by siamese neural network, the representation function learned by siamese neural network is denoted as $f_s(\cdot)$. The true label in the task is denoted as $y=0$ if the input pair have the same class labels and $y=1$ otherwise.  
The goal of SimCLR is to distinguish augmented views of one source image from views of other source images. Let $s_{i,j}=\boldsymbol{z}_i^{\top}\boldsymbol{z}_j/(\|\boldsymbol{z}_i\|\|\boldsymbol{z}_j\|)$ denote cosine similarity. The NT-Xent  ((normalized temperature-scaled cross-entropy)) loss is
\begin{equation}\label{eq:contrastive}
\mathcal{L}=\frac{1}{2N} \sum_{k=1}^{N} \left[\ell(2k-1,2k)+\ell(2k,2k-1)\right], \quad
\ell(i,j)=-\log \frac{\exp \left(s_{i,j} / \tau\right)}{\sum_{r=1}^{2 N} \mathbf{1}_{[r \neq i]} \exp \left(s_{i,r} / \tau\right)},
\end{equation}
where $\tau$ is the temperature controlling the hardness of negative samples. 
%which can be decomposed into two parts, uniformity loss and alignment loss. 
%which maximizes the similarity between the augmentated pairs, and the dissimilarity between augmentations generated from different data points.
%Various versions of contrastive loss can be generalized to this kind of form \cite{chen2021intriguing}. 
The first term inside $\ell(i,j)$ rewards alignment of a positive pair, whereas the log-sum-exp term distributes embeddings relative to the other views in the batch \citep{wang2020understanding, chen2021intriguing}.

\subsection{Feature suppression in self-supervised contrastive learning}
\label{subsec:fs-definition}
{\color{black}
We view a feature as a square-integrable function of an augmented input $X$. Let $U$ denote its source image, with population augmented-instance label $Y=U$. A scalar feature $h(X)$ is \emph{source-informative} if $U\not\!\perp h(X)$. A representation $f(X)$ is \emph{sufficient} for this task if $U\indep X\mid f(X)$.

For the SDR interpretation, assume that a smallest sufficient sub-$\sigma$-field $\mathcal G\subseteq\sigma(X)$ exists, with $U\indep X\mid\mathcal G$, and let $\mathcal C=L^2_0(\mathcal G)$ be its centered square-integrable function class. All $\sigma$-fields are interpreted modulo null sets. The class $\mathcal C$ can be infinite-dimensional even when $\mathcal G$ is generated by finitely many coordinates \citep{lee2013general}. Here $\sigma(X)$ is the $\sigma$-field generated by $X$, and $P_X$ and $P_U$ denote the marginal laws of $X$ and $U$. The space $L^2_0(P_X)$ consists of functions $h(X)$ with $\mathbb Eh(X)=0$ and $\mathbb E\{h(X)^2\}<\infty$; $L^2_0(P_U)$ is defined analogously. The subspace $L^2_0(\mathcal G)$ consists of the $\mathcal G$-measurable members of $L^2_0(P_X)$. Functions equal almost surely are identified.

For comparisons based on linear readout, fix a finite family of centered, source-informative target functions $h_1,\ldots,h_q\in\mathcal C$, specified independently of the representation being evaluated. For a square-integrable representation $f$, write
\[
\mathcal H(f)
=\operatorname{span}\{f_1-\mathbb Ef_1,\ldots,f_d-\mathbb Ef_d\}
\subset L^2_0(P_X).
\]
This closed, finite-dimensional span describes linear accessibility in the embedding.

\begin{definition}[Feature suppression relative to a target family]
\label{def:feature-suppression}
For the specified targets, define the linear-readout errors
\[
e_j(f)=\inf_{a\in\mathbb R^d}
\mathbb E\left[\{h_j(X)-a^\top(f(X)-\mathbb Ef(X))\}^2\right].
\]
The representation exhibits feature suppression relative to this family if $e_j(f)>0$ for at least one $j$, equivalently if at least one $h_j\notin\mathcal H(f)$.
\end{definition}

Linear accessibility and sufficiency describe different properties of a representation. A target can be measurable with respect to $\sigma(f)$ while having positive linear-readout error. We write $\sigma(f)=\sigma\{f(X)\}$ for the information generated by the representation.
}

\section{Fisher Contrastive Learning}\label{sec:fcl}
\subsection{FCL Formulation and Objective}
Feature suppression in SimCLR can degrade the performance of downstream tasks, especially
when some features are strong or easy to learn, so that they suppress other discriminant
features needed for prediction. We propose Fisher Contrastive Learning to address
this issue. FCL combines a contrastive objective with a Fisher discrimination
criterion and aims to learn representations that preserve the information needed to
distinguish the augmented classes. \fclnladd{Our analysis identifies when central projection improves the population FCL score and provides uniform linear-recovery bounds over a selected signal space. The main text states the assumptions and results; the supplement gives the proofs.} %. Such issues occur when some features are too easy or strong to learn other discriminant features for downstream tasks. 
%Even without strong augmentations, implicit regularization caused by over-parametrized deep networks can lead to the dimension-collapsing problem. 
\fclnladd{The population analysis below identifies the target of the two-view trace score in Eq.~\eqref{eq:empirical}.}
% In addition, our proposed algorithm can preserve the dimensionality of embeddings, further guaranteeing the embeddings are robust to feature suppression effects.
%Moreover, easy-to-learn shared features from augmented pairs can also lead to bad performance. In \cite{robinson2021can}, the lack of features learned in SimCLR can lead to bad downstream task performance. Sometimes, strong features suppress other features which is important to the learning of downstream tasks, thus, our goal is to preserve as much information as possible.

%As informed by the name, our proposed FCL leverages the idea of the FDA, which achieves Fisher consistency. %, which means the projected centroids are spread out as much as possible in terms of variance. This amounts to finding the principal component subspaces of the centroids themselves.

Because the contrastive loss in Eq.~(\ref{eq:contrastive}) distinguishes augmented views by their source image, we formulate SSCL as an augmented-instance classification task.
%We first generalize the goal of SSCL to the classification task, which aims to classify the augmentations from the same image to the same class, and augmentations from different images to different classes. 
%The augmented pairs from one image belong to the same class, whereas the augmentations from different images belong to different classes.
%For a dataset with $N$ images, we can get $2N$ images. %divided into $N$ classes, where each class has two data points. 
%The augmented pairs from one image belong to the same class, whereas the augmentations from different images belong to different classes.
The classification labels are defined such that the augmentations from the $j$-th data point belong to the $j$-th class. There are $N$ augmented instance classes in a batch containing $N$ original data points and their $2N$ augmented views. %Converting the SSCL problem to a classification task, we define the classification label as  $Y$, taking values from $\left\{1,...j,...,n\right\}$.
%The positive pairs augmented from image $j$ have the label  $Y=j$. 
%In each batch, 
%Various classifiers can be trained for the goal of self-supervised learning. 
%the goal of SSCL aligns with the goal of  FDA in the embedding space.
Fisher discriminant analysis identifies directions that separate classes by maximizing between-class variation relative to within-class variation. FCL applies this criterion after a neural-network mapping, allowing nonlinear discriminant functions of the original input \citep{dorfer2015deep, hastie1995penalized}.

In FCL, within-class scatter provides an analogue of positive-pair alignment, whereas between-class scatter measures separation among augmented instance classes.
For a minibatch containing $N$ original data points prior to augmentation, let
$\boldsymbol{z}_i^{(1)}=f(\tilde{\boldsymbol{x}}_{2i-1})$ and
$\boldsymbol{z}_i^{(2)}=f(\tilde{\boldsymbol{x}}_{2i})$ denote the two embeddings of the $i$th augmented instance class, and let
$\boldsymbol{\mu}_i=\{\boldsymbol{z}_i^{(1)}+\boldsymbol{z}_i^{(2)}\}/2$.
We define the normalized within-class scatter matrix as
\begin{equation}
\boldsymbol{\Sigma}_W
:=\frac{1}{2N}\sum_{i=1}^N\sum_{a=1}^2
\left(\boldsymbol{z}_i^{(a)}-\boldsymbol{\mu}_i\right)
\left(\boldsymbol{z}_i^{(a)}-\boldsymbol{\mu}_i\right)^{\top}
=\frac{1}{4N}\sum_{i=1}^N
\left(\boldsymbol{z}_i^{(1)}-\boldsymbol{z}_i^{(2)}\right)
\left(\boldsymbol{z}_i^{(1)}-\boldsymbol{z}_i^{(2)}\right)^{\top}.
\end{equation}
The matrix $\boldsymbol{\Sigma}_W$ is $d\times d$ and quantifies the contribution of within-class dispersion to the pooled covariance of the embeddings. Minimizing this scatter component encourages the two augmented views within each class to be close.
The normalized between-class scatter matrix measures how much each augmented-instance class mean deviates from the overall mean. It is defined as
%The within-class variance measures the variance within each class. The goal is to minimize within-class variance over between-class variance, ensuring that the members of each class are as close as possible to their respective class mean. The between-class variance of the embeddings measures how much each class mean differs from the overall mean,
\begin{equation}
\boldsymbol{\Sigma}_B
:=\frac{1}{N}\sum_{i=1}^N
\left(\boldsymbol{\mu}_{i}-\boldsymbol{\mu}\right)
\left(\boldsymbol{\mu}_{i}-\boldsymbol{\mu}\right)^{\top},
\end{equation}
where $\boldsymbol{\mu}=N^{-1}\sum_{i=1}^{N}\boldsymbol{\mu}_i$ is the overall mean embedding.
The normalized total scatter matrix of all $2N$ embeddings is therefore
\begin{equation}
\begin{split}
\boldsymbol{\Sigma}_T
&:=\frac{1}{2N}\sum_{i=1}^N\sum_{a=1}^2
\left(\boldsymbol{z}_i^{(a)}-\boldsymbol{\mu}\right)
\left(\boldsymbol{z}_i^{(a)}-\boldsymbol{\mu}\right)^{\top}\\
&=\boldsymbol{\Sigma}_W+\boldsymbol{\Sigma}_B.
\end{split}
\end{equation}
%where $c$ is the number of clusters, and $n_k$ is the number of samples in the $k$-th cluster. 
% \begin{equation}
% \tilde{\boldsymbol{S}}_B:=\sum_{k=1}^c \sum_{\ell=1, \ell \neq k}^c \sum_{i=1}^{n_k} \sum_{j=1}^{n_{\ell}}\left(f(\boldsymbol{x}_i^{(k)})-f(\boldsymbol{x}_j^{(\ell)})\right)\left(f(\boldsymbol{x}_i^{(k)})-f(\boldsymbol{x}_j^{(\ell)})\right)^{\top},
% \end{equation}
%where $n_l$ is the number of samples in the $l$-th cluster. 
The unregularized Fisher score is
\begin{equation}
\label{eq:obj}
\underset{f}{\operatorname{max}} \quad \Delta=  \operatorname{tr}\left(\boldsymbol{\Sigma}_W^{-1}\boldsymbol{\Sigma}_B\right),
\end{equation}
when $\boldsymbol{\Sigma}_W$ is nonsingular. The score is the sum of the generalized Fisher eigenvalues and increases when between-class separation is large relative to within-class dispersion \citep{fukunaga1990introduction}. The trace representation avoids computing the eigenvectors when only their summed eigenvalues are required.

% \begin{equation}
% \underset{\boldsymbol{U}}{\operatorname{max}} \:
%  \operatorname{tr}\left(\frac{\boldsymbol{U}^{\top} \boldsymbol{S}_B \boldsymbol{U}}{\boldsymbol{U}^{\top} \boldsymbol{S}_W U}\right),
% \end{equation}

Let $q\leq\min\{d,N-1,\operatorname{rank}(\boldsymbol{\Sigma}_W)\}$ be the number of Fisher directions under consideration and let $\boldsymbol{U}\in\mathbb{R}^{d\times q}$. On a subspace where $\boldsymbol{\Sigma}_W$ is positive definite, the corresponding constrained problem is
\begin{equation}
\underset{\boldsymbol{U}}{\operatorname{maximize}} \:  \operatorname{tr}\left(\boldsymbol{U}^{\top} \boldsymbol{\Sigma}_B \boldsymbol{U}\right), \qquad
\operatorname{subject\ to} \: \boldsymbol{U}^{\top} \boldsymbol{\Sigma}_W \boldsymbol{U}=\boldsymbol{I}_q.
\end{equation}
The columns of $\boldsymbol{U}$ define the Fisher discriminant directions. Because the constraint is matrix-valued, its Lagrange multiplier is initially a symmetric matrix $\boldsymbol{\Gamma}=\boldsymbol{\Gamma}^{\top}$. The corresponding Lagrangian is
\begin{equation}
\mathcal{L}(\boldsymbol{U},\boldsymbol{\Gamma})
=\operatorname{tr}\left(\boldsymbol{U}^{\top}\boldsymbol{\Sigma}_B\boldsymbol{U}\right)
-\operatorname{tr}\left\{\boldsymbol{\Gamma}\left(\boldsymbol{U}^{\top}\boldsymbol{\Sigma}_W\boldsymbol{U}-\boldsymbol{I}_q\right)\right\}.
\end{equation}
Since $\boldsymbol{\Sigma}_B$, $\boldsymbol{\Sigma}_W$, and $\boldsymbol{\Gamma}$ are symmetric, stationarity with respect to $\boldsymbol{U}$ gives
\begin{equation}
\boldsymbol{\Sigma}_B\boldsymbol{U}
=\boldsymbol{\Sigma}_W\boldsymbol{U}\boldsymbol{\Gamma}.
\end{equation}
Premultiplying by $\boldsymbol{U}^{\top}$ and using the constraint yields $\boldsymbol{\Gamma}=\boldsymbol{U}^{\top}\boldsymbol{\Sigma}_B\boldsymbol{U}$, which is symmetric. Moreover, the objective and constraint are invariant under the transformation $\boldsymbol{U}\mapsto\boldsymbol{U}\boldsymbol{Q}$ for any orthogonal matrix $\boldsymbol{Q}$. We may therefore choose $\boldsymbol{Q}$ to diagonalize $\boldsymbol{\Gamma}$. With $\widetilde{\boldsymbol{U}}=\boldsymbol{U}\boldsymbol{Q}$ and
\[
\boldsymbol{\Lambda}
=\boldsymbol{Q}^{\top}\boldsymbol{\Gamma}\boldsymbol{Q}
=\operatorname{diag}(\eta_1,\ldots,\eta_q),
\]
the stationarity equation becomes
\begin{equation}\label{eq:evd}
\boldsymbol{\Sigma}_B\widetilde{\boldsymbol{U}}
=\boldsymbol{\Sigma}_W\widetilde{\boldsymbol{U}}\boldsymbol{\Lambda}.
\end{equation}
\revblue{This choice of basis makes $\boldsymbol{\Lambda}$ diagonal. The columns of $\widetilde{\boldsymbol{U}}$ are generalized eigenvectors of the pair $(\boldsymbol{\Sigma}_B,\boldsymbol{\Sigma}_W)$, and $\boldsymbol{\Lambda}$ contains the corresponding generalized eigenvalues.}
For notational simplicity, we relabel $\widetilde{\boldsymbol{U}}$ as $\boldsymbol{U}$ below.

% The discrimination functions (a.k.a. canonical variates) in the embedding space are the 
% corresponding eigenvectors to separate positive samples from negative samples.
The columns $\boldsymbol{u}_j$ of $\boldsymbol{U}$ are linear Fisher directions in the embedding space; their compositions $\boldsymbol{u}_j^{\top}f(\cdot)$ are nonlinear discriminant functions of the original input \citep{chen2001generalization}.

We estimate $f$ with a neural network optimized by minibatch stochastic gradient descent. Direct differentiation through the generalized eigendecomposition in Eq.~(\ref{eq:evd}) can be numerically unstable.
% The discrimination functions in the embedding space are eigenvectors in the eigenvalue decomposition:
% \begin{equation}
%     \boldsymbol{S}_B \mathbf{e}_i=v_i\boldsymbol{S}_W\mathbf{e}_i,
% \end{equation}
% where $\mathbf{e}_i$ 
%For high-dimensional embedding space, $\tilde{\boldsymbol{S}}_W$ could be singular, thus, we add a regularization term to prevent the numeric instability, 
% \begin{equation}\label{eq:decomp_reg}
% \tilde{\boldsymbol{S}}_B \hat{\mathbf{e}}_i=\hat{v}_i\left(\tilde{\boldsymbol{S}}_W+\lambda \mathbf{I}\right) \hat{\mathbf{e}}_i,
% \end{equation}
% where $\hat{\mathbf{e}}_i$ is the regularized approximation of $\mathbf{e}_i$.:
%Note that the between-class variance $\boldsymbol{S}_B$ is positive semi-definite and all its eigenvalues are non-negative \cite{mardia1979multivariate}.
%When implemented,

To address this issue, we use a regularized trace criterion. With two views per augmented instance class, $\operatorname{rank}(\boldsymbol{\Sigma}_W)\leq\min\{N,d\}$, so singularity is structural whenever $d>N$ and may also occur otherwise. We therefore add a ridge term to the normalized within-class scatter matrix \citep{zhong2005rsir},
$\boldsymbol{\Sigma}_B \boldsymbol{U}=\left(\boldsymbol{\Sigma}_W+\lambda_{\mathrm{F}}\boldsymbol{I}_d\right) \boldsymbol{U} \boldsymbol{\Lambda}$,
where $\lambda_{\mathrm{F}}>0$ is a ridge hyperparameter selected by grid search. The empirical loss function is
\begin{equation}
\label{eq:empirical}
\mathcal{L}= - \log \  \operatorname{tr}\left(\left(\boldsymbol{\Sigma}_W+\lambda_{\mathrm{F}}\boldsymbol{I}_d\right)^{-1}\boldsymbol{\Sigma}_B\right).
\end{equation}

We optimize this objective using minibatch stochastic gradient descent.
% To train the neural network, we use batch training to speed up the training process instead of using the full batch. 
%We anayze the computational complexity for training each batch of size $n$.

{\color{black}
\subsection{Population Target and Finite-Batch Theory}
{
We analyze the population target and finite-batch approximation of the trace criterion in Eq.~\eqref{eq:empirical}. Let $(U_i,X_i^{(1)},X_i^{(2)})$ be independent and identically distributed source--view triples. Conditional on the source $U_i$, the two augmented views $X_i^{(1)}$ and $X_i^{(2)}$ are independent draws from the same conditional augmentation distribution $P_{X\mid U=U_i}$. For a fixed measurable encoder $f:\mathcal X\to\mathbb R^d$, assume $\mathbb E\|f(X)\|^2<\infty$, finite $d$, and a fixed $\lambda=\lambda_{\mathrm F}>0$. Define
\[
W_f=\mathbb E\{\operatorname{Cov}(f(X)\mid U)\},\qquad
B_f=\operatorname{Cov}\{\mathbb E(f(X)\mid U)\}.
\]

\begin{theorem}[Two-view trace limit]\label{thm:main-trace-limit}
For $N\geq2$, the empirical scatter matrices defined above satisfy
\begin{equation}\label{eq:two-view-moments}
\mathbb E\boldsymbol{\Sigma}_W=W_f/2,\qquad
\mathbb E\boldsymbol{\Sigma}_B=\frac{N-1}{N}(B_f+W_f/2).
\end{equation}
Define the empirical score and its population counterpart by
\begin{equation}\label{eq:main-raw-limit}
\begin{aligned}
J_N(f)
&:=\operatorname{tr}\{(\boldsymbol{\Sigma}_W+\lambda I_d)^{-1}\boldsymbol{\Sigma}_B\},\\
J(f)
&:=\operatorname{tr}\{(W_f/2+\lambda I_d)^{-1}(B_f+W_f/2)\}.
\end{aligned}
\end{equation}
Then, for this fixed encoder, $J_N(f)\to J(f)$ almost surely as $N\to\infty$.
\end{theorem}
\theoremexplain{Theorem~\ref{thm:main-trace-limit} identifies the population Fisher score approximated by the two-view minibatch criterion. It accounts for within-source variability in both scatter components and supplies the common objective for the subsequent finite-batch, central-projection, and signal-recovery results. Supplement Section~1 gives the proof.}
% The empirical score in Eq.~\eqref{eq:main-raw-limit} is the quantity inside the logarithm in Eq.~\eqref{eq:empirical}. Its population numerator contains both between-source variation $B_f$ and within-source variation $W_f/2$.

% The proof uses the conditional covariance of the view average and difference, the strong law, and continuity of the ridge inverse. Supplement Section~1 gives the details and a deterministic perturbation bound.

}

\begin{proposition}[Finite-batch approximation of the trace loss]\label{prop:main-batch-risk}
Suppose additionally that $\|f(X)\|\leq R$ almost surely, and put
\[
K_{d,R,\lambda}=3\sqrt d\,R^2/\lambda+R^4/\lambda^2.
\]
Then
\[
\mathbb E|J_N(f)-J(f)|
\leq K_{d,R,\lambda}/\sqrt N.
\]
For fixed $\varepsilon>0$, define
$\mathcal R_{N,\varepsilon}(f)=\mathbb E[-\log\{J_N(f)+\varepsilon\}]$
and
$\ell_\varepsilon(f)=-\log\{J(f)+\varepsilon\}$.
Over any deterministic encoder class $\mathcal F$ with common $R,d,\lambda$,
\[
\sup_{f\in\mathcal F}|\mathcal R_{N,\varepsilon}(f)-\ell_\varepsilon(f)|
\leq K_{d,R,\lambda}/(\varepsilon\sqrt N).
\]
Here $\eta_N\geq0$ is the allowed excess of the expected-batch risk above its infimum over $\mathcal F$.
Any $\eta_N$-approximate global minimizer $f_N$ of $\mathcal R_{N,\varepsilon}$ therefore satisfies
\[
\ell_\varepsilon(f_N)-\inf_{f\in\mathcal F}\ell_\varepsilon(f)
\leq 2K_{d,R,\lambda}/(\varepsilon\sqrt N)+\eta_N.
\]
\end{proposition}
Proposition~\ref{prop:main-batch-risk} controls the expected score approximation error at rate $N^{-1/2}$ and separates finite-batch error from optimization error in the population log-loss bound. On the normalized representation class of Theorem~\ref{main-nl-recovery}, this loss gap determines a Fisher-score deficit, connecting minibatch objective approximation to the subsequent uniform signal-recovery bound. Supplement Section~1 gives the proof.

\textbf{Central projection under SDR.}
The central $\sigma$-field $\mathcal G$ is the smallest sufficient
$\sigma$-field of $X$ for $U$: after conditioning on $\mathcal G$,
the remaining variation in $X$ carries no additional information
about the source $U$. Its centered, square-integrable functions form
the central function class $\mathcal C$. The operator
$Th=\mathbb E(h\mid U)$ describes the mean of a feature $h$ for each
source. We regard $T$ as a map from $L^2_0(P_X)$ to $L^2_0(P_U)$, with inner product $\langle h,k\rangle=\mathbb E(hk)$ in each space. Its adjoint $T^*$ satisfies $\langle Th,a\rangle=\langle h,T^*a\rangle$ and is given by $T^*a=\mathbb E\{a(U)\mid X\}$. Here $\ker T=\{h:Th=0\}$ is the null space, and $\mathcal C^\perp$ consists of functions orthogonal to every member of $\mathcal C$. The composition $T^*T$ applies $T$ first; $\operatorname{Ran}(T^*T)=\{T^*Th:h\in L^2_0(P_X)\}$ is its range. The overline denotes closure under mean-square convergence in $L^2(P_X)$. Completeness means that a central feature whose
source-conditional mean is zero must itself be zero:
$h\in L^2_0(\mathcal G)$ and $\mathbb E(h\mid U)=0$ imply $h=0$.
Under sufficiency and completeness, Supplement Section~1 proves
\[
\ker T=\mathcal C^\perp,\qquad
\overline{\operatorname{Ran}(T^*T)}=\mathcal C,
\qquad \mathcal C=L^2_0(\mathcal G).
\]
These identities characterize the central function class through
source-conditional averaging. They connect the SDR formulation to
FCL because the between-source matrix
$B_f=\operatorname{Cov}\{\mathbb E(f\mid U)\}$ summarizes variation
in these source-conditional means.

{\color{black}
The following theorem connects the SDR projection to the population FCL criterion. It uses sufficiency; completeness enters the preceding operator identification.

\begin{theorem}[Central projection and the population FCL criterion]
\label{thm:main-central-projection}
Suppose $\mathcal G\subseteq\sigma(X)$ satisfies $U\indep X\mid\mathcal G$, let $f:\mathcal X\to\mathbb R^d$ satisfy $\mathbb E\|f(X)\|^2<\infty$, and fix $\lambda>0$. Define
\[
p=\mathbb E(f\mid\mathcal G),\qquad r=f-p.
\]
Then
\begin{equation}\label{eq:main-central-moments}
\mathbb E(p\mid U)=\mathbb E(f\mid U),\qquad
B_p=B_f,\qquad W_f=W_p+\mathbb E(rr^\top).
\end{equation}
For the population score in Eq.~\eqref{eq:main-raw-limit},
\begin{equation}\label{eq:main-central-score}
B_f\succeq\lambda I_d
\quad\Longrightarrow\quad
J(p)\geq J(f).
\end{equation}
The inequality is strict if $B_f-\lambda I_d$ is positive definite and $\mathbb E\|r\|^2>0$.
\end{theorem}
Theorem~\ref{thm:main-central-projection} shows that central projection preserves source-conditional means while reducing noncentral variation. Under the stated signal condition, this improves or preserves the Fisher score, giving the objective an SDR interpretation. Theorem~\ref{main-nl-recovery} then relates score accuracy to linear recovery of selected signals. Supplement Section~1 gives the proof and a quantitative extension.
% Central projection preserves the current representation's source-conditional means and reduces its average within-source variation. The signal condition in Theorem~\ref{thm:main-central-projection} requires the between-source variance in every unit direction to be at least $\lambda$. Under this condition, the projection also yields an equally large or larger population FCL score.

% The condition reflects the contribution of within-source variation to the numerator. For a scalar source-independent feature with variance $v>0$ and $B_f=0$,
% \[
% J=\frac{v}{v+2\lambda}>0.
% \]
% A positive score can therefore arise from within-source variation alone. Supplement Section~1 gives the proof and a quantitative version: with bounded outputs, a stronger signal margin, and an encoder class closed under central projection, a population score deficit bounds the remaining noncentral variation. The next subsection establishes a uniform linear-recovery bound over a selected signal space.
}

{\color{black}
\subsection{Signal Recovery with a Nonlinear Encoder and Linear Head}
\label{main-nl-subsection}
We next connect the population FCL score to uniform linear recovery
over a selected signal space when the encoder and projection head
are learned jointly. In this subsection, $\theta$ denotes the encoder parameters and $p$ denotes its output dimension. Let $f_\theta:\mathcal X\to\mathbb R^p$ be a measurable
encoder with finite second moments, and write
\begin{equation}
z_{\theta,A}(X)=A\{f_\theta(X)-\mathbb E f_\theta(X)\},
\qquad A\in\mathbb R^{d\times p},\quad 1\leq d\leq p.
\label{main-nl-architecture}
\end{equation}
Both $\theta$ and $A$ may vary. Impose the population constraint
\begin{equation}
\operatorname{Cov}(z_{\theta,A})=I_d,
\quad\text{equivalently}\quad
A\operatorname{Cov}\{f_\theta(X)\}A^\top=I_d.
\label{main-nl-normalization}
\end{equation}
\revgreen{The constraint couples the head with the encoder covariance through $A\operatorname{Cov}\{f_\theta(X)\}A^\top=I_d$. Its validity depends on both components of the representation.}

Let $\mathcal V$ be a fixed closed subspace of $L^2_0(P_X)$
containing every candidate output coordinate, with inner product
$\langle h,k\rangle=\mathbb E(hk)$ and norm $\|h\|_2^2=\mathbb E(h^2)$.
One may take $\mathcal V=L^2_0(P_X)$. Define
\begin{equation}
T_{\mathcal V}:\mathcal V\to L^2_0(P_U),\qquad
T_{\mathcal V}h=\mathbb E(h\mid U),\qquad
\mathcal K_{\mathcal V}=T_{\mathcal V}^*T_{\mathcal V}.
\label{main-nl-operator}
\end{equation}
The adjoint is taken for this fixed domain. Explicitly, $T_{\mathcal V}^*a=P_{\mathcal V}\mathbb E\{a(U)\mid X\}$, where $P_{\mathcal V}$ is orthogonal projection onto $\mathcal V$. The operator
$\mathcal K_{\mathcal V}$ is a positive self-adjoint contraction
determined by the source--view distribution, even as $\theta$ changes.

Assume there are orthonormal functions $\phi_1,\ldots,\phi_d\in\mathcal V$
and constants satisfying
\begin{equation}
\mathcal K_{\mathcal V}\phi_i=\mu_i\phi_i,\qquad
1\geq\mu_1\geq\cdots\geq\mu_d>\nu\geq0.
\label{main-nl-eigenfunctions}
\end{equation}
For $S_\star=\operatorname{span}\{\phi_1,\ldots,\phi_d\}$, {write $S_\star^\perp$ for its orthogonal complement in $L^2_0(P_X)$ and assume
\begin{equation}
\langle h,\mathcal K_{\mathcal V}h\rangle\leq\nu\|h\|_2^2
\quad\text{for every }h\in\mathcal V\cap S_\star^\perp,
\qquad\gamma=\mu_d-\nu>0.
\label{main-nl-gap}
\end{equation}
 \fclnladd{The gap separates the chosen signal space from its orthogonal
complement in the fixed ambient space $\mathcal V$.} \fclnladd{Let $\Phi=(\phi_1,\ldots,\phi_d)^\top$. Consider the target class
\begin{equation}\label{main-nl-target-class}
\mathcal H_\star
=\bigcup_{q=1}^{\infty}
\left\{C\Phi:C\in\mathbb R^{q\times d},\ \|C\|_{\mathrm{op}}\leq1\right\}.
\end{equation}
Here $d$ is the dimension of the selected signal space and the final representation, $q$ is the finite number of target coordinates, and each row of $C$ specifies a linear combination of the eigenfeatures. The operator-norm constraint controls the target scale. The signal space and target class are fixed independently of the representation being evaluated.}

Let $\mathfrak Z$ be a nonempty class of outputs
\eqref{main-nl-architecture} satisfying the normalization and
$\mathcal V$-membership conditions. Use the original population score
\begin{equation}
J(z)=\operatorname{tr}\!\left\{
(W_z/2+\lambda I_d)^{-1}(B_z+W_z/2)\right\},
\qquad\lambda>0,
\label{main-nl-score}
\end{equation}
with the same between-source and within-source covariances as above.
Define
\begin{equation}
g_\lambda(t)=\frac{1+t}{1+2\lambda-t},\qquad
m_\lambda=\frac{2(1+\lambda)}{(1+2\lambda)^2},\qquad
J_{\mathcal V}^\star=\sum_{i=1}^d g_\lambda(\mu_i),
\label{main-nl-ideal-score}
\end{equation}
and the class optimum, approximation deficit, and optimization deficit
\begin{align}
J_{\mathfrak Z}^\star
&=\sup_{v\in\mathfrak Z}J(v),\notag\\
a_{\mathfrak Z}
&=J_{\mathcal V}^\star-J_{\mathfrak Z}^\star,\qquad
\eta(z)=J_{\mathfrak Z}^\star-J(z).
\label{main-nl-deficits}
\end{align}
\fclnladd{For $\boldsymbol h\in\mathcal H_\star$ with $q$ coordinates, define the target errors from the head and encoder by
}\begin{align}
\mathcal E_{\boldsymbol h}(z)
&=\inf_{D\in\mathbb R^{q\times d}}
\mathbb E\|\boldsymbol h-Dz\|_2^2,\notag\\
\mathcal E_{\boldsymbol h}(f_\theta)
&=\inf_{B\in\mathbb R^{q\times p}}
\mathbb E\|\boldsymbol h-B\{f_\theta-\mathbb E f_\theta\}\|_2^2.
\label{main-nl-readout}
\end{align}
\fclnladd{Write $L(z)=\mathcal E_\Phi(z)$ for the total squared error in recovering the $d$ eigenfeatures.}

\begin{theorem}[\fclnladd{Uniform signal recovery under joint optimization}]
\label{main-nl-recovery}
Under \eqref{main-nl-normalization}--\eqref{main-nl-gap} and the
class and target conditions above, $J_{\mathcal V}^\star$ is the
maximum score over all orthonormal $d$-tuples in $\mathcal V$.
For every admissible pair $(\theta,A)$, the deficits in
\eqref{main-nl-deficits} are nonnegative and
\fclnladd{\begin{equation}\label{main-nl-error-bound}
\begin{aligned}
\sup_{\boldsymbol h\in\mathcal H_\star}\mathcal E_{\boldsymbol h}(f_\theta)
&\leq\sup_{\boldsymbol h\in\mathcal H_\star}\mathcal E_{\boldsymbol h}(z_{\theta,A})
=L(z_{\theta,A})\\
&\leq\min\left\{d,\frac{a_{\mathfrak Z}+\eta(z_{\theta,A})}{m_\lambda\gamma}\right\}.
\end{aligned}
\end{equation}}
If some admissible pair realizes $z=Q\Phi$ for an orthogonal
$d\times d$ matrix $Q$, then $a_{\mathfrak Z}=0$, the class maximum
is attained, and every joint global maximizer has this form.
\fclnladd{Every such maximizer therefore permits simultaneous exact linear recovery of every $\boldsymbol h\in\mathcal H_\star$ from both its projected representation and its encoder output.}
\end{theorem}
Theorem~\ref{main-nl-recovery} converts architecture and optimization deficits into a uniform linear-recovery bound over $\mathcal H_\star$, with sensitivity determined by the spectral gap. The bound applies to both encoder and projected representations. Under the stated conditions, a realizable global optimum recovers every target exactly, while approximate solutions have controlled recovery error. Supplement Section~5 gives the proof.
% \fclnladd{The uniform statement follows from the same residual covariance for every target. Put $R=\mathbb E(\Phi z^\top)$ and $M_z=I_d-RR^\top\succeq0$. Then
% \[
% \mathcal E_{C\Phi}(z)=\operatorname{tr}(CM_zC^\top)
% \leq\operatorname{tr}(M_z)=L(z),
% \]
% because $C^\top C\preceq I_d$. Equality is attained at $q=d$ and $C=I_d$, corresponding to recovery of $\Phi$ itself. Thus one bound controls all finite target dimensions and all allowed coefficient matrices in Eq.~\eqref{main-nl-target-class}.

% If $S_\star\subseteq\mathcal C$, the bound simultaneously controls linear feature suppression for the nonzero scalar targets in $\mathcal H_\star$ under Definition~\ref{def:feature-suppression}. Exact recovery holds at a realizable global optimum; a positive near-optimal error still counts as suppression under that definition. When $\mathcal V=L^2_0(P_X)$ and a central sufficient $\sigma$-field $\mathcal G$ exists, the positive-eigenvalue functions belong to $\mathcal C=L^2_0(\mathcal G)$. The guarantee concerns linear combinations in this selected spectral space; the cell-image application evaluates the resulting representations through downstream prediction.} \fclnladd{The theorem analyzes joint nonlinear population optimization under
% an additional identity-covariance constraint. }
% } Supplement Section~5 gives the full proof, an explicit bound on
% $a_{\mathfrak Z}$ from approximate eigenfeatures, and
% counterexamples showing why the stated conditions are needed.

\subsection{Computational Cost}
\revgreen{For a minibatch containing $N$ original samples, the two augmented views yield $2N$ embeddings in $\mathbb{R}^d$. Excluding the encoder computation shared by both methods, forming the FCL scatter matrices requires $O(Nd^2)$ operations. Let $\boldsymbol{C}\in\mathbb{R}^{N\times d}$ have row $i$ equal to $(\boldsymbol{\mu}_i-\boldsymbol{\mu})^{\top}/\sqrt{N}$, so that $\boldsymbol{\Sigma}_B=\boldsymbol{C}^{\top}\boldsymbol{C}$, and write $\boldsymbol{\Sigma}_W+\lambda_{\mathrm{F}}\boldsymbol{I}_d=\boldsymbol{L}\boldsymbol{L}^{\top}$, where $\boldsymbol{L}$ is a Cholesky factor. The trace in Eq.~\eqref{eq:empirical} equals $\|\boldsymbol{L}^{-1}\boldsymbol{C}^{\top}\|_F^2$. This computation uses a triangular solve and requires $O(Nd^2+d^3)$ time and $O(Nd+d^2)$ auxiliary memory per minibatch. A standard dense SimCLR implementation forms pairwise similarities among the $2N$ views, requiring $O(N^2d)$ time and $O(Nd+N^2)$ auxiliary memory, with constants associated with the two views suppressed. FCL can be cheaper when $N\gg d$; the direct $d\times d$ factorization can dominate when $d\gg N$. When $d=\Theta(N)$, the losses have the same asymptotic time and memory orders. In the $d\gg N$ regime, the low-rank scatter structure also permits FCL evaluation in $O(dN^2+N^3)$ time and $O(Nd+N^2)$ memory. These rates apply to the trace loss in Eq.~\eqref{eq:empirical}; end-to-end cost also includes encoder passes, implementation, hardware, and distributed communication.}

\section{Simulated Experiments}\label{sec:sim}

%We present three simulation studies designed to illustrate feature suppression in
% self-supervised contrastive learning (SSCL) and to evaluate the proposed Fisher
% Contrastive Learning (FCL) method under controlled conditions. The first study uses a
% low-dimensional Gaussian example with known class-relevant and class-irrelevant
% directions. The second and third studies, Random Bits and Digits on Flowers
% \citep{chen2021intriguing}, create high-dimensional settings where strong nuisance
% signals can dominate the representation. Together, these examples complement the
% white blood cell application by isolating the mechanisms through which feature
% suppression arises and by demonstrating how FCL mitigates them.
% \subsection{Gaussian Distribution with feature suppression effects}
% We present simulated experiments showing the performance of FCL. We adapt a classification task introduced by \cite{meng2020sufficient}  to a self-supervised learning task. The example includes two C-shaped curves (i.e.,  two classes) with random Gaussian noise in a two-dimensional subspace embedded in $\mathbb{R}^{18}$ as shown in panel (a) of Figure \ref{fig:sdr}.

\subsection{Gaussian Distribution with Feature Suppression}
\label{subsec:gaussian-sim}

Our first study adapts a classification task introduced by
\citet{meng2020sufficient} to the SSCL setting. We compare the class
separation and downstream classification accuracy of representations
learned by FCL and SimCLR in a controlled setting where additional
noise coordinates may hinder representation learning.

We generate two classes forming C-shaped curves in the first two
coordinates of $\mathbb{R}^{18}$ and add 16 independent Gaussian noise
coordinates. For each class, we generate 5,000 observations with the
first two coordinates defined by
\[
\begin{aligned}
\text{Class 1:} \quad &
X_1 = 20 \cos(\theta) + Z_1 + 5,\quad
X_2 = 10 \sin(\theta) + Z_2 - 5, \\
\text{Class 2:} \quad &
X_1 = -20 \cos(\theta) - Z_1 - 5,\quad
X_2 = 10 \sin(\theta) + Z_2 + 5,
\end{aligned}
\]
where $Z_1,Z_2\sim N(0,1)$ and
$\theta\sim N\big(\pi,(0.25\pi)^2\big)$ are mutually independent.
The remaining coordinates $X_3,\ldots,X_{18}$ are independently
generated from $N(0,5)$, independently of $(X_1,X_2)$, with the same
distribution in both classes.
Panel~(a) of Figure~\ref{fig:sdr} displays the 10,000 observations
in the first two coordinates.

We train a single-hidden-layer encoder under the SSCL framework. Data augmentation is
implemented by adding Gaussian noise $N(0,1)$ to each input, which generates positive
pairs for contrastive learning. The embedding dimension is fixed at $d=10$. We compare
the standard SimCLR objective with the proposed FCL objective, using the same encoder
architecture and training settings. After pretraining, we use the learned embeddings to
fit a $k$-nearest neighbor classifier ($k=5$) and evaluate downstream classification
accuracy.

The results for the reported run are summarized in Figure~\ref{fig:sdr}. Panels (b) and (c) show the learned embedding
spaces after linear transformation by Fisher discriminant directions. Using the
representation learned by FCL, a $k$-nearest neighbor classifier achieves an accuracy of
85.30\%. In contrast, SimCLR yields an accuracy of 58.00\%, which is close to random
guessing in this two-class setting. In panel~(c), the SimCLR representation shows little separation of the two classes after the Fisher projection. Together with the lower $k$-nearest-neighbor accuracy, this pattern is consistent with suppression of the informative $(X_1,X_2)$ structure. 

\revgreen{The FCL representation in panel~(b) provides clearer class separation after the Fisher projection and yields higher downstream accuracy in the reported run.}

\begin{figure}[htbp]
\centering
\includegraphics[width=0.9\textwidth]{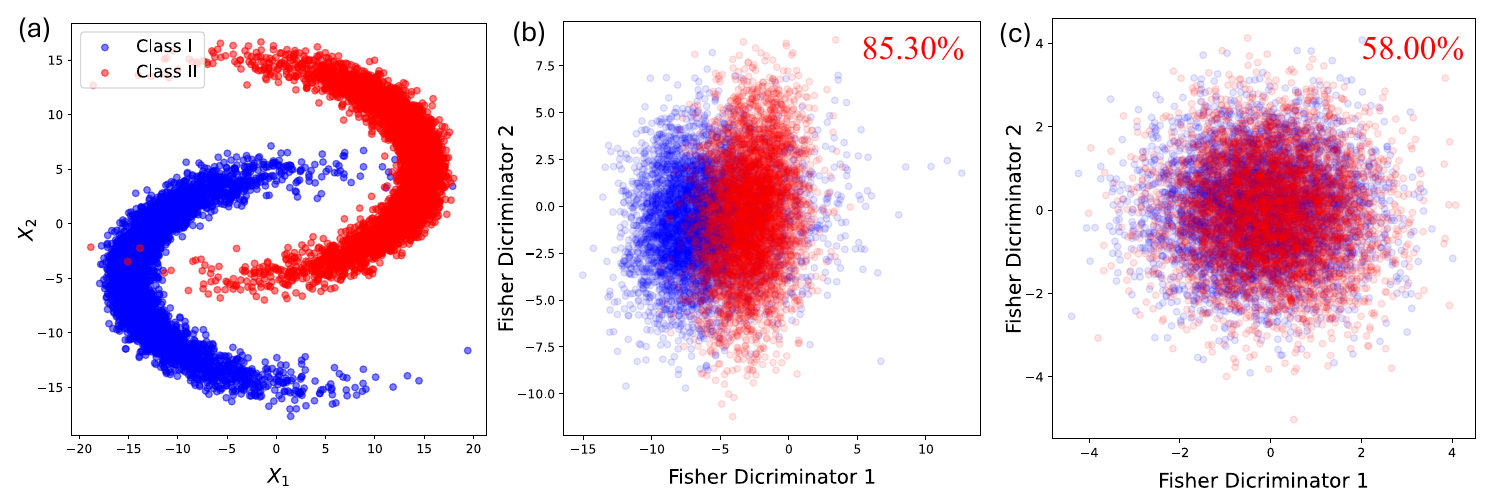}
\caption{Controlled feature-suppression simulation. (a) The two informative input coordinates. (b,c) Fisher projections of the embeddings learned by FCL and SimCLR, respectively. A 5-nearest-neighbor classifier fitted to the reported FCL and SimCLR embeddings achieves accuracies of 85.30\% and 58.00\%, respectively.}\label{fig:sdr}

\end{figure}

\begin{table}[htp]
\centering
\caption{\revblue{Point estimates of accuracy for cell-type classification using pretrained representations.}}
\label{tab:cell_type}
\small
\begin{tabular}{lrrrrrr}
\hline
\textbf{Task} & \textbf{SimCLR} & \textbf{FCL} & \added{\textbf{VICReg}} & \added{\textbf{SimSiam}} & \textbf{BYOL} & \added{\textbf{DINO}} \\
\hline
Cell Type & 98.45\% & \added{\textbf{99.42\%}} & \added{97.87\%} & \added{98.64\%} & \added{98.64\%} & \added{98.84\%} \\
\hline
\end{tabular}
\end{table}

% \begin{figure}[htbp]
% \centering
% \includegraphics[width=1\linewidth]{figs/Umap_FCL_SimCLR.pdf}
% \vspace{-10pt}
% \caption{UMAP visualization of the learned representation by SimCLR (panels a and b) and FCL (panels c and d). Panels a and c show the UMAP of representations of the training dataset, and panels b and d show the UMAP of representations of the testing dataset. The blue points are non-APL cells, and the orange points are APL cells.} 
% \label{fig:umap_APL}
% \end{figure}

\section{Application to Acute Promyelocytic Leukemia Diagnosis}\label{sec:exp}
Acute promyelocytic leukemia (APL) is a rare but treatable subtype of acute myeloid leukemia (AML) and constitutes a hematologic emergency: delays in diagnosis can lead to rapid, fatal bleeding, whereas early recognition followed by targeted therapy yields an excellent prognosis \citep{sidhom2021deep}. In many clinical settings, particularly where access to molecular testing (e.g., PCR) is limited, peripheral blood smear morphology is used to triage suspected APL cases. The application studied here distinguishes patients with APL from patients with other AML subtypes in a small single-center dataset.

\revblue{We pretrain WBC representations with FCL on the large public dataset described in Section~\ref{subsec:wbc_data}. We then transfer these representations to the smaller APL dataset through the two-phase framework in Figure~\ref{fig:workflow}.}%

Phase~1 performs self-supervised representation learning on the large WBC dataset, with a ResNet encoder serving as $f(\cdot)$. \revblue{The model is trained using image augmentations without manual supervision. We compare FCL with SimCLR, VICReg \citep{bardes2022vicreg}, SimSiam \citep{chen2021exploring}, BYOL \citep{grill2020bootstrap}, and DINO \citep{caron2021emerging}. Table~\ref{tab:cell_type} reports cell-type classification accuracy using the pretrained representations.}

\revblue{Phase~2 initializes the encoder with weights from Phase~1 and adds a supervised classification head for cells sampled from patients with APL or other AML subtypes. The discovery cohort contains 82 patients (22 APL and 60 other AML), and the held-out prospective cohort contains 24 patients (12 APL and 12 other AML). We fine-tune the pretrained model on the discovery cohort and evaluate it on the prospective cohort. Patients define the cohort split, and each cell inherits its source patient's molecular diagnosis.}

\revblue{For evaluation, the model produces an APL probability for each available cell. We average these probabilities within each patient, classify the patient using the resulting mean probability, and calculate accuracy, AUROC, and AUPRC across the 24 patients in the prospective cohort.}

\revblue{FCL achieves the highest point estimates of all three metrics among the compared methods (Table~\ref{tab:APL_result}), which include MLP as a benchmark.}

% \added{Table~\ref{tab:APL_cell_result} reports the corresponding cell-level predictions before patient aggregation. VICReg also has the highest cell accuracy (0.7270), AUROC (0.7529), and AP (0.7629) among the four added methods. These metrics measure agreement with inherited patient diagnoses, not expert-labeled disease status of individual cells. Cells within a patient are dependent, so the 24-patient evaluation remains the primary clinical endpoint. No new FCL cell-level estimate is substituted for the original manuscript results.}

\revblue{FCL organizes the representation by comparing between-instance variation with augmentation-induced within-instance variation. We hypothesize that this geometry makes task-relevant information more accessible to a supervised classifier trained with limited patient-level labels.}

\revblue{The patient-level point estimates provide preliminary evidence for the utility of FCL representations in diagnostic transfer. Our proposed method motivates further evaluation across various settings outside the WBC microscopic images.}

\begin{table}[htbp]
\centering

\caption{Patient-level point estimates for classification of APL versus other AML in the prospective cohort ($n=24$: 12 APL and 12 other AML). }
\begin{tabular}{c c c c } 
 \hline
 & Accuracy & AUROC & AUPRC\\ 
\hline
MLP & 0.5833 & 0.7014 &  0.7494\\
SimCLR & 0.6667 & 0.7153 &  0.7263\\
FCL & \textbf{0.7500}  & \textbf{0.8264}& \textbf{0.8405} \\
\added{VICReg} & \added{0.7083} & \added{0.7986} & \added{0.8332} \\
\added{SimSiam} & \added{0.6667} & \added{0.7014} & \added{0.7260} \\
\added{BYOL} & \added{0.6667} & \added{0.7292} & \added{0.7445} \\
\added{DINO} & \added{0.6667} & \added{0.7222} & \added{0.6654} \\
\hline
\end{tabular} 
\label{tab:APL_result}
\end{table}

\section{Conclusions}
\label{sec:con}
We investigated feature suppression in self-supervised contrastive learning (SSCL) and its impact on automated diagnosis of acute promyelocytic leukemia from white blood cell images. 
\fclnladd{For the two-view Fisher trace criterion, we proved fixed-encoder population limits and finite-batch approximation bounds. Under the stated SDR assumptions, completeness identifies the closure of a conditional-expectation operator range. A main-text theorem shows that central projection improves the original population FCL score under an explicit between-source signal condition.} \fclnladd{For nonlinear encoders with linear heads and covariance-normalized outputs, we bound the linear-recovery error uniformly over normalized linear combinations in a selected leading eigenspace. The bound separates architectural approximation and population optimization deficits and depends on the spectral gap separating that space from the remaining directions.}
%Motivated by this perspective, we proposed Fisher Contrastive Learning, which reinterprets SSCL as a nonlinear dimension reduction problem and maximizes a Fisher discrimination criterion to preserve class-relevant features in representation.

\revblue{In the Gaussian simulation with known signal and noise structure, the reported FCL embeddings showed clearer class separation and higher downstream accuracy than SimCLR.}

\revblue{Our APL case study used self-supervised pretraining on a large WBC corpus followed by supervised fine-tuning on the labeled discovery cohort. FCL achieved the highest point estimates of accuracy, AUROC, and AUPRC among the compared methods on the prospective cohort. These results provide preliminary evidence for the utility of FCL representations in diagnostic learning with limited labeled patients.}

\revblue{The proposed framework connects SSCL to nonlinear dimension reduction.} Future work includes extending FCL to other data modalities and architectures, and developing uncertainty quantification and model assessment tools for downstream tasks based on FCL representations.
% In this paper, we propose a novel contrastive learning method, FCL, to address the feature suppression effects prevalent in SSCL.  Feature suppression effects, where easy-to-learn and class-irrelevant features suppress other class-relevant features, is a common problem in SSCL. 
% Our method offers a robust solution to feature suppression effects with theoretical guarantees, retaining the central sufficient dimension reduction space.
% We demonstrate the advantages of our method using image datasets that exhibit feature suppression effects, as well as benchmark image datasets.
% Compared to other self-supervised learning methods, our proposed method is more robust to various levels of feature suppression effects.
% When the assumption of completeness of the central dimension reduction class is not satisfied, the proposed algorithm is still unbiased but no longer exhaustive. To recover a function class that is larger than the current one, we can use the idea of the sliced average variance estimator \citep{lee2013general} when the variance of the distribution of various classes differs in self-supervised learning. 
% Additionally, our proposed framework has broader applications in multi-modal data fusion. Instead of
% ``CLIP''ing \citep{radford2021learning} two modalities like images and text using contrastive loss, the proposed framework can integrate multiple modalities using Fisher contrastive loss. 

% \begin{center}
% {\large\bf SUPPLEMENTARY MATERIAL}
% \end{center}

\begin{acks}[Acknowledgments]
Correspondence: Ping Ma (pingma@uga.edu) and Wenxuan Zhong (wenxuan@uga.edu). 
\end{acks}

\begin{supplement}
\stitle{Supplement to ``Fisher Contrastive Learning to Mitigate Feature Suppression: A Case Study in Acute Promyelocytic Leukemia Diagnosis''}
\sdescription{\revblue{\fclnladd{\fclnladd{The supplement contains proofs of the two-view trace limit, finite-batch approximation bounds, completeness and central-projection results, and uniform recovery bounds for nonlinear encoders with linear heads. It also provides implementation details and analytical comparisons of FCL with VICReg and Fisher-style discriminative objectives.}}}}
\end{supplement}

} % Close the existing blue revision group before the supplement.

% The supplement is incorporated below; retain its section numbers 1--5.
\clearpage
\section*{Supplementary material}
\revblue{\fclnladd{\fclnladd{This supplement gives population limits and finite-batch bounds for the two-view Fisher trace criterion and proves the completeness and central-projection results under explicit sufficient-dimension-reduction assumptions.}} It also provides implementation details and analytical comparisons with VICReg and prior Fisher-style discriminative objectives.} \fclnladd{\fclnladd{Section~5 establishes uniform recovery guarantees over normalized linear combinations in a selected signal space for jointly optimized nonlinear encoders and linear projection heads, with complete proofs, approximation bounds, and examples under explicit population normalization and spectral assumptions.}}

\setcounter{section}{0}
\setcounter{subsection}{0}
\setcounter{equation}{0}
\setcounter{theorem}{0}
\renewcommand{\theequation}{S\arabic{equation}}
\renewcommand{\thetheorem}{S\thesection.\arabic{theorem}}
\ifcsname theHsection\endcsname
  \renewcommand{\theHsection}{supp.\arabic{section}}
\fi
\ifcsname theHequation\endcsname
  \renewcommand{\theHequation}{supp.\arabic{equation}}
\fi
% BEGIN INCORPORATED SUPPLEMENT BODY

{\color{black}
\section{Theory for the Two-View Fisher Trace Criterion}\label{app:thm}
{\color{black}
\revgreen{This section establishes fixed-encoder population limits and finite-batch bounds for the regularized trace score under the two-view sampling model. It then introduces the central $\sigma$-field and derives conditional-expectation and projection results under the stated SDR assumptions.}
}

\subsection{Sampling model and population target}
\begin{assumption}[Two-view sampling]\label{ass:two-view-sampling}
{\color{black}
The source--view triples $(U_i,X_i^{(1)},X_i^{(2)})$, $i\geq1$, are independent and identically distributed. Conditional on $U_i$, the two views are independent draws from the same conditional augmentation distribution $P_{X\mid U_i}$. The source and view spaces are standard Borel spaces. The measurable encoder $f:\mathcal X\to\mathbb R^d$ is fixed independently of these triples, $d<\infty$, $\mathbb E\|f(X)\|^2<\infty$, and the ridge parameter $\lambda>0$ is fixed.
}
\end{assumption}
\revgreen{Let $U$ denote the population source variable and write $Z=f(X)$. Define}
\begin{equation}\label{eq:supp-pop-moments}
W_f=\mathbb E\{\operatorname{Cov}(Z\mid U)\},\qquad
B_f=\operatorname{Cov}\{\mathbb E(Z\mid U)\},\qquad
H_f=W_f/2,\qquad C_f=B_f+H_f.
\end{equation}
For $Z_i^{(a)}=f(X_i^{(a)})$, let
\[
M_i=(Z_i^{(1)}+Z_i^{(2)})/2,\quad
D_i=(Z_i^{(1)}-Z_i^{(2)})/2,\quad
\bar M_N=N^{-1}\sum_{i=1}^N M_i.
\]
The main-text scatter matrices are exactly
\begin{equation}\label{eq:supp-emp-moments}
H_N=N^{-1}\sum_{i=1}^N D_iD_i^\top=\boldsymbol{\Sigma}_W,\qquad
S_N=N^{-1}\sum_{i=1}^N M_iM_i^\top-\bar M_N\bar M_N^\top
=\boldsymbol{\Sigma}_B.
\end{equation}
{\color{black}
The empirical and population scores are
\begin{align}
J_N(f)&=\operatorname{tr}\{(H_N+\lambda I_d)^{-1}S_N\},
\notag\\
J(f)&=\operatorname{tr}\{(H_f+\lambda I_d)^{-1}C_f\}.
\label{eq:supp-raw-score}
\end{align}
The empirical score is the quantity inside the main-text logarithmic loss.

\begin{theorem}[Two-view moments and score limit]\label{thm:two-view-limit}
Under Assumption~\ref{ass:two-view-sampling},
\[
\mathbb EH_N=H_f,\qquad
\mathbb ES_N=\frac{N-1}{N}C_f.
\]
As $N\to\infty$, almost surely,
\[
H_N\to H_f,\qquad S_N\to C_f,\qquad
J_N(f)\to J(f).
\]
For every sample, the scores satisfy the deterministic bound
\begin{equation}\label{eq:supp-raw-perturb}
|J_N-J|
\leq \lambda^{-1}\|S_N-C_f\|_*
 +\lambda^{-2}\operatorname{tr}(C_f)\|H_N-H_f\|_{\mathrm{op}}.
\end{equation}
Here $\|\cdot\|_*$ and $\|\cdot\|_{\mathrm{op}}$ denote nuclear and operator norms.
\end{theorem}

\begin{proof}
Conditional independence gives
$\mathbb E(D_iD_i^\top\mid U_i)=\operatorname{Cov}(Z\mid U_i)/2$ and
$\operatorname{Cov}(M_i\mid U_i)=\operatorname{Cov}(Z\mid U_i)/2$.
Also, $\mathbb E(M_i\mid U_i)=\mathbb E(Z\mid U_i)$, so the law of total covariance gives $\operatorname{Cov}(M_i)=C_f$. The expectation of the covariance with divisor $N$ is $(N-1)C_f/N$, proving both moment identities.

The second-moment assumption makes every entry of $D_iD_i^\top$ and $M_iM_i^\top$ integrable. The strong law applied to these entries and to $M_i$ therefore gives the two matrix limits. All eigenvalues of $H_N+\lambda I_d$ are at least $\lambda$; inversion is continuous on this set, giving the score limit.

For the perturbation bound, put $A_N=H_N+\lambda I_d$ and $A=H_f+\lambda I_d$. Then
\[
\operatorname{tr}(A_N^{-1}S_N-A^{-1}C_f)
=\operatorname{tr}\{A_N^{-1}(S_N-C_f)\}
 +\operatorname{tr}\{(A_N^{-1}-A^{-1})C_f\}.
\]
Trace duality bounds the first term by $\lambda^{-1}\|S_N-C_f\|_*$. The identity
$A_N^{-1}-A^{-1}=A_N^{-1}(A-A_N)A^{-1}$
and $C_f\succeq0$ bound the second by
$\lambda^{-2}\|H_N-H_f\|_{\mathrm{op}}\operatorname{tr}(C_f)$.
\end{proof}
}

\subsection{Finite-batch score and expected-loss control}
\begin{proposition}[Finite-batch approximation]\label{prop:batch-risk}
In addition to Assumption~\ref{ass:two-view-sampling}, suppose $\|f(X)\|\leq R$ almost surely. Define
\[
K_{d,R,\lambda}=\frac{3\sqrt d\,R^2}{\lambda}+\frac{R^4}{\lambda^2}.
\]
Then
\begin{equation}\label{eq:supp-mean-score-bound}
\mathbb E|J_N(f)-J(f)|
\leq \frac{K_{d,R,\lambda}}{\sqrt N}.
\end{equation}
For any fixed $\varepsilon>0$, let
\[
\ell_{N,\varepsilon}(f)=-\log\{J_N(f)+\varepsilon\},\qquad
\ell_{\varepsilon}(f)=-\log\{J(f)+\varepsilon\},\qquad
\mathcal R_{N,\varepsilon}(f)=\mathbb E\ell_{N,\varepsilon}(f).
\]
If every encoder in a deterministic class $\mathcal F$ satisfies the same bound $R$, with common $d$ and $\lambda$, then
\begin{equation}\label{eq:supp-risk-bound}
\sup_{f\in\mathcal F}
|\mathcal R_{N,\varepsilon}(f)-\ell_\varepsilon(f)|
\leq \frac{K_{d,R,\lambda}}{\varepsilon\sqrt N}.
\end{equation}
Consequently, any $\eta_N$-approximate global minimizer $f_N\in\mathcal F$ of the expected batch loss, meaning
$\mathcal R_{N,\varepsilon}(f_N)\leq
\inf_{f\in\mathcal F}\mathcal R_{N,\varepsilon}(f)+\eta_N$,
satisfies
\begin{equation}\label{eq:supp-risk-excess}
\ell_\varepsilon(f_N)-\inf_{f\in\mathcal F}\ell_\varepsilon(f)
\leq \frac{2K_{d,R,\lambda}}{\varepsilon\sqrt N}+\eta_N.
\end{equation}
\end{proposition}

\begin{proof}
Set $m=\mathbb EM_i$, $Q=\mathbb E(M_iM_i^\top)$, and $Q_N=N^{-1}\sum_iM_iM_i^\top$. Since $\|M_i\|,\|D_i\|\leq R$, the variance identity for independent Hilbert-space averages and Jensen's inequality give
\[
\mathbb E\|H_N-H_f\|_F\leq R^2/\sqrt N,\qquad
\mathbb E\|Q_N-Q\|_F\leq R^2/\sqrt N,\qquad
\mathbb E\|\bar M_N-m\|\leq R/\sqrt N.
\]
For vectors $a,b$, $\|aa^\top-bb^\top\|_F\leq(\|a\|+\|b\|)\|a-b\|$. Therefore
$\mathbb E\|S_N-C_f\|_F\leq3R^2/\sqrt N$.
Also $\operatorname{tr}(C_f)=\mathbb E\|M_i-m\|^2\leq R^2$.
Apply \eqref{eq:supp-raw-perturb}, $\|A\|_*\leq\sqrt d\|A\|_F$, and $\|A\|_{\mathrm{op}}\leq\|A\|_F$ to obtain \eqref{eq:supp-mean-score-bound}.

Both raw scores are nonnegative, and $t\mapsto-\log(t+\varepsilon)$ is $1/\varepsilon$-Lipschitz on $[0,\infty)$. This gives \eqref{eq:supp-risk-bound} with the same bound for every $f$. Writing $a_N=K_{d,R,\lambda}/(\varepsilon\sqrt N)$,
\[
\ell_\varepsilon(f_N)\leq\mathcal R_{N,\varepsilon}(f_N)+a_N
\leq\inf_{f\in\mathcal F}\mathcal R_{N,\varepsilon}(f)+\eta_N+a_N
\leq\inf_{f\in\mathcal F}\ell_\varepsilon(f)+\eta_N+2a_N,
\]
which proves \eqref{eq:supp-risk-excess} without requiring the infimum to be attained.
\end{proof}

\revgreen{Equation~\eqref{eq:supp-risk-bound} compares deterministic expected-batch risks, with the supremum outside the expectation. Under independent fresh-batch sampling, $\mathcal R_{N,\varepsilon}$ averages the logarithmic loss over batches. The model treats each encoder as a fixed map applied separately to every view and assumes a common output bound and fixed ridge. Application to fitted networks also involves sample reuse, batch-dependent transformations, and optimization error.}

\revgreen{For the unfloored loss in the main text, $J(f)>0$ gives almost-sure convergence of $-\log J_N(f)$ on the eventually positive scores. Expected-loss control additionally requires control near zero. For example, if both scores are at least $a>0$ almost surely for every encoder under consideration, the bound in \eqref{eq:supp-risk-bound} applies to the unfloored loss with $a$ in place of $\varepsilon$. A zero-score batch with positive probability makes the expected unfloored loss infinite even when the population score is positive.}

\subsection{Central $\sigma$-fields and completeness}
SDR expresses source sufficiency as $U\indep X\mid\mathcal G$ \citep{li1991sliced,cook2002dimension,li2018sufficient,lee2013general}. All $\sigma$-fields and equalities of random variables below are understood modulo $P$-null sets. Write $L^2_0(P_X)$ for centered square-integrable functions of $X$.

\begin{assumption}[Central $\sigma$-field]\label{ass:central-sigma}
There exists a smallest sub-$\sigma$-field $\mathcal G\subseteq\sigma(X)$ for which $U\indep X\mid\mathcal G$. Define the central function class
\[
\mathcal C=L^2_0(\mathcal G)\subseteq L^2_0(P_X).
\]
For results explicitly invoking completeness, assume that
\[
h\in\mathcal C,\quad \mathbb E(h\mid U)=0
\quad\Longrightarrow\quad h=0\ \text{almost surely}.
\]
\end{assumption}
\revgreen{Existence of the smallest sufficient $\sigma$-field is an assumption here; dominated-family conditions yielding its existence are given by \citet{lee2013general}. The class $\mathcal C$ contains all centered square-integrable $\mathcal G$-measurable functions, including nonlinear functions of any generating coordinates. It can therefore be infinite-dimensional even when $\mathcal G=\sigma\{g_1(X),\ldots,g_r(X)\}$ for finitely many coordinates. Finite-dimensional $L^2_0(\mathcal G)$ restricts $\mathcal G$, modulo null sets, to finitely many atoms.}

Define the bounded conditional-expectation operator
\[
T:L^2_0(P_X)\to L^2_0(P_U),\qquad Th=\mathbb E(h\mid U),
\]
and the orthogonal projection $P_{\mathcal G}h=\mathbb E(h\mid\mathcal G)$. Its adjoint is $T^*a=\mathbb E\{a(U)\mid X\}$.

\begin{proposition}[What completeness identifies]\label{prop:complete-range}
Under Assumption~\ref{ass:central-sigma} including completeness,
\begin{equation}\label{eq:supp-complete-range}
\ker T=\mathcal C^\perp,\qquad
\overline{\operatorname{Ran}T^*}
=\overline{\operatorname{Ran}(T^*T)}=\mathcal C.
\end{equation}
The closures and orthogonal complement are taken in $L^2_0(P_X)$.
\end{proposition}
\begin{proof}
Conditional independence gives
$\mathbb E(h\mid U,\mathcal G)=\mathbb E(h\mid\mathcal G)$, hence $T=TP_{\mathcal G}$. Thus $\mathcal C^\perp=\ker P_{\mathcal G}\subseteq\ker T$. If $Th=0$, then $T(P_{\mathcal G}h)=0$. Since $P_{\mathcal G}h\in\mathcal C$, completeness implies $P_{\mathcal G}h=0$, proving the reverse inclusion. The Hilbert-space identities
$\overline{\operatorname{Ran}T^*}=(\ker T)^\perp$
and
$\overline{\operatorname{Ran}(T^*T)}=(\ker(T^*T))^\perp$,
together with $\ker(T^*T)=\ker T$, prove \eqref{eq:supp-complete-range}.
\end{proof}
\revgreen{Building on the completeness-based SDR characterization in \citet{lee2013general}, Proposition~\ref{prop:complete-range} identifies the central function class through the closure of a conditional-expectation operator range. Completeness yields the kernel equality. If $U$ has $K$ possible values, injectivity of $T$ on $\mathcal C$ gives $\dim\mathcal C\leq K-1$, since $T$ maps into $L^2_0(P_U)$. Full-class completeness is thus a substantive condition in finite-source models. Theorem~\ref{thm:two-view-limit} and Proposition~\ref{prop:batch-risk} use the sampling and moment assumptions stated for those results.}

{\color{black}
\subsection{Projection property and its limits}
For a square-integrable vector encoder, write
$p(X)=\mathbb E\{f(X)\mid\mathcal G\}$,
$r=f-p$, and $V_r=\mathbb E(rr^\top)$.
The following theorem concerns the original population score in \eqref{eq:supp-raw-score}.

\begin{theorem}[Central projection of the original trace score]\label{thm:central-projection}
Fix $\lambda>0$. Under the sufficiency part of Assumption~\ref{ass:central-sigma}, without requiring completeness,
\begin{equation}\label{eq:supp-projection-moments}
B_f=B_p,\qquad W_f=W_p+V_r.
\end{equation}
Consequently,
\[
B_f\succeq\lambda I_d\quad\Longrightarrow\quad
J(p)\geq J(f).
\]
The inequality is strict if $B_f-\lambda I_d$ is positive definite and $V_r\neq0$.

If additionally $\|f(X)\|\leq R$ and $B_f\succeq(\lambda+b)I_d$ for some $b>0$, then
\begin{equation}\label{eq:supp-projection-gap}
J(p)-J(f)
\geq \frac{b\,\mathbb E\|r\|^2}{2(\lambda+R^2/2)^2}.
\end{equation}
Under these additional conditions, if an admissible class $\mathcal F$ is closed under $f\mapsto p$ and $f\in\mathcal F$ is within $\eta$ of its global supremum for $J$, then
\[
\mathbb E\|f-\mathbb E(f\mid\mathcal G)\|^2
\leq \frac{2(\lambda+R^2/2)^2}{b}\eta.
\]
\end{theorem}

\begin{proof}
Since $\mathbb E(r\mid\mathcal G)=0$, we have $\mathbb Er=0$ and $\operatorname{Cov}(p,r)=0$. Conditional independence further gives $\mathbb E(r\mid U)=0$, as in the proof of Proposition~\ref{prop:complete-range}, without the completeness step. Thus $\mathbb E(f\mid U)=\mathbb E(p\mid U)$ and $B_f=B_p$. Apply the total covariance identity to
$\operatorname{Cov}(f)=\operatorname{Cov}(p)+V_r$
to obtain $W_f=W_p+V_r$.

Put $A_0=W_p/2+\lambda I_d$ and $A_1=A_0+V_r/2=W_f/2+\lambda I_d$. Then $A_0^{-1}-A_1^{-1}\succeq0$. The exact identity
\begin{equation}\label{eq:supp-raw-rewrite}
J(f)
=d+\operatorname{tr}\{(W_f/2+\lambda I_d)^{-1}(B_f-\lambda I_d)\}
\end{equation}
gives
\[
J(p)-J(f)
=\operatorname{tr}\{(A_0^{-1}-A_1^{-1})(B_f-\lambda I_d)\}.
\]
This is nonnegative when $B_f\succeq\lambda I_d$.

For strictness and the quantitative bound, set $A_t=A_0+tV_r/2$. Differentiating the inverse and integrating gives
\[
A_0^{-1}-A_1^{-1}
=\frac12\int_0^1 A_t^{-1}V_rA_t^{-1}\,dt.
\]
Under the output bound, $W_f\preceq R^2I_d$, so $A_t\preceq(\lambda+R^2/2)I_d$. If $B_f\succeq(\lambda+b)I_d$, then $B_f-\lambda I_d\succeq bI_d$, and cyclicity of the trace yields
\[
\begin{aligned}
\operatorname{tr}\{(B_f-\lambda I_d)A_t^{-1}V_rA_t^{-1}\}
&=\operatorname{tr}\{V_r^{1/2}A_t^{-1}(B_f-\lambda I_d)A_t^{-1}V_r^{1/2}\}\\
&\geq \frac{b\,\operatorname{tr}(V_r)}{(\lambda+R^2/2)^2}.
\end{aligned}
\]
Integration proves \eqref{eq:supp-projection-gap}. Without the output bound, positive definiteness of $B_f-\lambda I_d$ still makes this trace positive whenever $V_r\neq0$, proving strictness. Finally, projection closure implies that the score gap between $p$ and $f$ is at most $\eta$, proving the last claim.
\end{proof}

\revgreen{The following scalar example illustrates the role of the signal condition. Let a centered encoder be independent of $U$, with $\operatorname{Var}\{f(X)\}=v>0$. Then $B_f=0$, $W_f=v$, and}
\[
J(f)=\frac{v}{v+2\lambda}>0.
\]
For example, if $X$ and $U$ are independent, the central $\sigma$-field is trivial and $p=0$, giving $J(p)=0$. Thus the raw score can reward purely within-source noise.

\revgreen{Under the assumptions of Theorem~\ref{thm:central-projection}, central projection removes noncentral variation while preserving source-conditional means. The quantitative conclusion also uses projection closure of the encoder class and the stated signal margin. For a $\mathcal G$-measurable representation, source sufficiency is equivalent to $\sigma(f)=\mathcal G$; a general representation preserves the central information when $\mathcal G\subseteq\sigma(f)$. Section~\ref{fclnl-supp:section} addresses linear recovery of a selected signal space under population covariance normalization and spectral separation.}
}
}

\section{Implementation Details}\label{app:imple}
% \subsection{Modified Optimization Target}
% For a minibatch of $N$ augmented instance classes and embedding dimension $d$, let $q=\min(d,N-1)$ and let $\hat{v}_1,\ldots,\hat{v}_q$ denote the regularized generalized Fisher eigenvalues. The main-text empirical trace criterion weights all $q$ eigenvalues equally through their sum. An optional Deep-LDA-style modification instead emphasizes eigenvalues near the smallest one \citep{dorfer2015deep}. For a margin $m>0$, define
% \begin{equation}
% \Theta_m=\left\{j\in\{1,\ldots,q\}:\hat{v}_j<\min_{1\leq \ell\leq q}\hat{v}_\ell+m\right\}
% \end{equation}
% and maximize
% \begin{equation}
% J_m=\frac{1}{|\Theta_m|}\sum_{j\in\Theta_m}\hat{v}_j.
% \end{equation}
% This margin-selected eigenvalue objective is not algebraically identical to the trace loss and requires a generalized eigendecomposition. The theoretical comparisons in the remainder of this supplement concern the trace criterion unless the margin modification is named explicitly. \revblue{The score limits and finite-batch bounds in Section~\ref{app:thm} concern the trace specification, not this margin-selected objective or a loss with additional scale or entropy penalties.}
\subsection{Experiment Details}
All methods use ImageNet-initialized ResNet50 backbones and are evaluated after 200 epochs of WBC pretraining. FCL uses batch size 256 and Adam with a constant learning rate of $10^{-3}$. The remaining methods use batch size 256: SimCLR, VICReg, and BYOL use LARS with peak learning rates of 0.15, 0.1, and 0.1, respectively, whereas SimSiam and DINO use SGD with peak learning rates of 0.025 and 0.015.  WBC evaluation uses 6,162 images retained from the official training split and 1,030 official validation images. Images are resized to $224\times224$ and normalized using training-set statistics. A class-balanced logistic-regression probe ($C=1$, L-BFGS, maximum 2,000 iterations, tolerance $10^{-4}$) is fitted to standardized, frozen 2,048-dimensional pre-projector features, with BatchNorm frozen and feature standardization fitted only on the training set.  A 1,024-image subset of the validation split is also used for pretraining monitoring.

\subsection{Patient prediction and model selection}
The Discovery cohort contains 82 patients (22 APL and 60 other AML) and 12,127 retained cell images; the prospective cohort contains 24 patients (12 APL and 12 other AML) and 3,465 retained cell images. We use the Discovery cohort for training and model selection and the prospective cohort for evaluation, with APL as the positive class. Each patient's score is the arithmetic mean of the predicted APL probabilities across that patient's retained cells.

All methods use full-encoder fine-tuning for 10 epochs with a $2048\to256\to1$ MLP head, ReLU activation, and dropout probability 0.2. Training minimizes binary cross-entropy using cell labels inherited from each patient's diagnosis, with Adam, a learning rate of $10^{-4}$, weight decay of $10^{-4}$, and batch size 32. These methods use a fixed patient cutoff of 0.5, with scores strictly above the cutoff classified as APL. The baseline comparator uses a randomly initialized ResNet50 and the same MLP head, trained end to end with the same optimizer settings. 

%\subsubsection{Unsupervised pre-training}

%For benchmark datasets, we follow the settings in SimCLR to design the architecture. The backbone network is ResNet-18 and the projector is a two layers MLP after ResNet-18's global average pooling layer ($pool_5$). The input and output dimensions of each layer are 128, and each layer is followed by ReLU activation function.

%The unsupervised pre-training uses base learning rate=0.03 with a cosion dacay schedule for 600 epochs, weight decay=0.0005, momentum=0.9, warm up epochs=10, batch size=128 with a SGD opeimizer.

%\subsubsection{Linear evaluation}

%For benchmark datasets, after pre-training the network, we freeze the parameters of it and train a supervised linear classifier. The features used for training are from ResNet-18's global average pooling layer ($pool_5$). The linear classifier training uses base learning rate=30 with a cosine decay schedule for 30 epochs, momentum=0.9, batch size=256 with a SGD optimizer.

{\color{black}
\section{Relationship to VICReg}\label{app:fcl-vicreg}

VICReg combines pair alignment with penalties on low coordinate-wise standard deviations and off-diagonal branch covariances \citep{bardes2022vicreg}. We relate these terms to the empirical scatter matrices used by FCL. For the two embeddings $\boldsymbol z_i^{(1)}$ and $\boldsymbol z_i^{(2)}$ of instance $i$, define $\boldsymbol\delta_i=\boldsymbol z_i^{(1)}-\boldsymbol z_i^{(2)}$, $\boldsymbol\mu_i=(\boldsymbol z_i^{(1)}+\boldsymbol z_i^{(2)})/2$, and $\boldsymbol m_i=\boldsymbol\mu_i-N^{-1}\sum_r\boldsymbol\mu_r$. Then
\begin{equation}
\boldsymbol\Sigma_W=\frac{1}{4N}\sum_i\boldsymbol\delta_i\boldsymbol\delta_i^\top,
\qquad
\boldsymbol\Sigma_B=\frac{1}{N}\sum_i\boldsymbol m_i\boldsymbol m_i^\top,
\qquad
J_{\mathrm{FCL}}=\operatorname{tr}\{(\boldsymbol\Sigma_W+\lambda_{\mathrm F}\boldsymbol I_d)^{-1}\boldsymbol\Sigma_B\},
\label{eq:fcl-vicreg-fcl-loss}
\end{equation}
where $\lambda_{\mathrm F}>0$, and the trace loss is $-\log J_{\mathrm{FCL}}$ on positive scores.

Let $\boldsymbol C_a$ be the sample covariance of branch $a$, with denominator $N-1$. In a convention that sums the alignment error over embedding coordinates, VICReg has the form
\begin{align}
\mathcal L_{\mathrm{VICReg}}
={}&\frac{\alpha}{N}\sum_i\|\boldsymbol z_i^{(1)}-\boldsymbol z_i^{(2)}\|^2
+\frac{\beta}{d}\sum_{a=1}^2\sum_j[\gamma-\sqrt{(\boldsymbol C_a)_{jj}+\epsilon}]_+
\nonumber\\
&+\frac{\eta}{d}\sum_{a=1}^2\|\operatorname{off}(\boldsymbol C_a)\|_{\mathrm F}^2,
\label{eq:vicreg-full-loss}
\end{align}
where $\alpha,\beta,\eta>0$, $\gamma>0$, $\epsilon>0$, $[t]_+=\max(t,0)$, and $\operatorname{off}$ removes the diagonal. Averaging the alignment error over coordinates rescales $\alpha$ by $1/d$. Its unweighted value is exactly $4\operatorname{tr}(\boldsymbol\Sigma_W)$.

The branch covariances also depend on between-instance scatter. Write $\overline{\boldsymbol\delta}=N^{-1}\sum_i\boldsymbol\delta_i$, $\boldsymbol e_i=\boldsymbol\delta_i-\overline{\boldsymbol\delta}$, and $\boldsymbol K=N^{-1}\sum_i(\boldsymbol m_i\boldsymbol e_i^\top+\boldsymbol e_i\boldsymbol m_i^\top)$. Expanding the centered branch embeddings gives
\begin{equation}
\boldsymbol C_a=\frac{N}{N-1}\left\{
\boldsymbol\Sigma_W+\boldsymbol\Sigma_B
-\frac14\overline{\boldsymbol\delta}\,\overline{\boldsymbol\delta}^{\top}
+\frac{(-1)^{a+1}}{2}\boldsymbol K\right\},\qquad a=1,2.
\label{eq:vicreg-branch-decomposition}
\end{equation}
Thus VICReg regularizes branch covariances containing both scatter components and finite-batch cross terms. When $\overline{\boldsymbol\delta}=0$ and $\boldsymbol K=0$, both covariances equal $N(\boldsymbol\Sigma_W+\boldsymbol\Sigma_B)/(N-1)$. FCL uses the full within-scatter matrix to weight the between-instance scatter. If the two scatter matrices are simultaneously diagonal, this weighting becomes
\begin{equation}
J_{\mathrm{FCL}}=\sum_{j=1}^d\frac{b_j}{w_j+\lambda_{\mathrm F}}.
\label{eq:fcl-directionwise}
\end{equation}
At the population level, the empirical between-instance scatter converges to $B_f+W_f/2$ under the two-view model; source-conditional variability therefore enters both parts of the population FCL score in Eq.~\eqref{eq:main-raw-limit}.

The two objectives also have a useful relationship on the covariance-normalized population class of Section~\ref{fclnl-supp:section}. With $\operatorname{Cov}(z)=I_d$, VICReg's population variance penalty is constant and its off-diagonal penalty is zero. Conditional independence of the views gives
\begin{equation}
\mathbb E\|z(X^{(1)})-z(X^{(2)})\|^2
=2\{d-\operatorname{tr}(B_z)\},
\qquad B_z=\operatorname{Cov}\{\mathbb E[z(X)\mid U]\}.
\label{eq:vicreg-normalized-population}
\end{equation}
Under the spectral gap and realizability conditions stated there, both objectives select the same leading signal space. This comparison evaluates VICReg's penalties at the population covariance and imposes the same covariance constraint on both objectives. Their original training losses differ in the weighting of suboptimal representations and in their dependence on output scale.

For a constant representation, $J_{\mathrm{FCL}}=0$; appending a zero output coordinate leaves the score unchanged. VICReg assigns a positive variance penalty to a constant coordinate when $\gamma>\sqrt\epsilon$. Under perfect alignment, $J_{\mathrm{FCL}}=\lambda_{\mathrm F}^{-1}\operatorname{tr}(\boldsymbol\Sigma_B)$, which can grow without bound with unrestricted scale. These properties distinguish the trace objective's behavior from the covariance-constrained recovery statement.

\section{Relationship to Fisher-Style Discriminative Objectives}\label{app:fcl-ghojogh}

Ghojogh et al.\ introduced Fisher Discriminant Triplet (FDT) and Fisher Discriminant Contrastive (FDC) losses for Siamese networks \citep{ghojogh2020fisher}. Let $\boldsymbol z=\boldsymbol P_{\mathrm G}^{\top}\boldsymbol o$ denote their final linear embedding. Their positive and negative latent scatter matrices sum anchor--neighbor and anchor--distant outer products, respectively, and include diagonal enrichments $\tau_{+,\mathrm G}I$ and $\tau_{-,\mathrm G}I$. For the corresponding projected traces $T_+$ and $T_-$, the losses are
\begin{align}
\mathcal L_{\mathrm{FDT}}&=[(2-\rho_{\mathrm G})T_+-\rho_{\mathrm G}T_-+\alpha_{\mathrm G}]_+,
\label{eq:ghojogh-fdt}\\
\mathcal L_{\mathrm{FDC}}&=(2-\rho_{\mathrm G})\widetilde T_+
+[-\rho_{\mathrm G}\widetilde T_-+\alpha_{\mathrm G}]_+,
\label{eq:ghojogh-fdc}
\end{align}
where $0<\rho_{\mathrm G}<1$, $\alpha_{\mathrm G}>0$, and the tildes denote scatters formed from the FDC pair sets. Each projected trace is a sum of squared embedding distances plus $\tau_{\pm,\mathrm G}\|\boldsymbol P_{\mathrm G}\|_{\mathrm F}^2$. These objectives combine aggregate distances with fixed coefficients and the displayed hinge placements.

A common augmented-instance construction makes the relation to FCL explicit. Define the average positive-pair scatter $\overline{\boldsymbol G}_+=N^{-1}\sum_i\boldsymbol\delta_i\boldsymbol\delta_i^\top$ and the exhaustive negative-view scatter
\begin{equation}
\overline{\boldsymbol G}_-
=\frac{1}{4N(N-1)}\sum_{i\ne j}\sum_{a,c=1}^2
(\boldsymbol z_i^{(a)}-\boldsymbol z_j^{(c)})
(\boldsymbol z_i^{(a)}-\boldsymbol z_j^{(c)})^\top.
\label{eq:ghojogh-all-negative}
\end{equation}
Expansion around the instance centroids yields
\begin{equation}
\overline{\boldsymbol G}_+=4\boldsymbol\Sigma_W,
\qquad
\overline{\boldsymbol G}_-=2\boldsymbol\Sigma_W+\frac{2N}{N-1}\boldsymbol\Sigma_B.
\label{eq:ghojogh-negative-decomp}
\end{equation}
Consequently,
\begin{equation}
J_{\mathrm{FCL}}=\frac{N-1}{2N}\operatorname{tr}\!\left\{
\left(\frac14\overline{\boldsymbol G}_++\lambda_{\mathrm F}\boldsymbol I_d\right)^{-1}
\left(\overline{\boldsymbol G}_--\frac12\overline{\boldsymbol G}_+\right)\right\}.
\label{eq:ghojogh-fcl-pair-representation}
\end{equation}
The positive-pair term removes the within-instance contribution from the negative-view scatter, leaving a between-centroid component weighted by a regularized matrix inverse. FDT and FDC apply scalar coefficients to aggregate traces. In the simultaneous diagonal setting of Eq.~\eqref{eq:fcl-directionwise}, their distance components can agree for representations with the same $\sum_jw_j$ and $\sum_jb_j$, while FCL distinguishes their directional allocations. This comparison uses exhaustive, branch-symmetric pair sets and separates the diagonal-enrichment terms. Sampled pair sets introduce an additional dependence on the sampling rule; separately averaging positive and negative sums also changes their relative weights.

In the original experiments, labels or histopathology metadata determine positive and negative relations. Augmented views can also define these relations for FDT and FDC. FCL combines augmented-instance centroids with directional Fisher normalization. The diagonal enrichments above act through the final layer norm, whereas the FCL ridge enters the within-scatter inverse. A fixed ridge also makes the FCL score depend on output scale. Sections~\ref{app:thm} and~\ref{fclnl-supp:section} characterize its population target, finite-batch approximation, and conditional signal recovery.
}

% Audited nonlinear recovery addition: append immediately before the supplement bibliography.
% All displayed additions are red. No new macros, theorem environments, or packages are required.
{\color{black}
\section{Signal Recovery with Nonlinear Encoders and Linear Projection Heads}
\label{fclnl-supp:section}

\fclnladd{This section connects the original population FCL score to uniform
linear recovery over a selected nonlinear signal space. The encoder and
projection head may be optimized jointly. \fclnladd{\fclnladd{The results use identity covariance of the final output and
spectral separation. Exact recovery further requires realizability
of the selected signal space; the approximate bound includes the
joint architecture's approximation deficit.}}}

We use $\mathcal V$ for a fixed ambient function space and
$\boldsymbol h$ for the target vector. The operator $T_{\mathcal V}$
is the restriction to $\mathcal V$ of the conditional-expectation
operator $T$ in Section~\ref{app:thm}. The score $J_\lambda(z)$ below
is the population raw score $J(z)$ in
\eqref{eq:supp-raw-score}, under the additional covariance constraint.

\subsection{Nonlinear encoder, linear head, and the original FCL score}
\label{fclnl-supp:sec:architecture}

Let \((U,X)\) denote a source and its augmented view. The input
space may be nonlinear or non-Euclidean. Let
\(f_\theta:\mathcal X\to\mathbb R^p\) be a measurable nonlinear encoder
with finite second moments, and set
\begin{equation}
 \widetilde f_\theta(X)=f_\theta(X)-\mathbb E f_\theta(X),\qquad
 z_{\theta,A}(X)=A\widetilde f_\theta(X),\qquad
 A\in\mathbb R^{d\times p},\quad 1\le d\le p.
 \label{fclnl-supp:eq:architecture}
\end{equation}
Both \(\theta\) and \(A\) may vary during population optimization.
\fclnladd{Centering specifies the covariance and linear-readout convention.}

The normalization is imposed on the final representation:
\begin{equation}
 \mathbb E\{z_{\theta,A}z_{\theta,A}^\top\}=I_d,
 \quad\text{equivalently}\quad
 A\Sigma_\theta A^\top=I_d,
 \qquad \Sigma_\theta=\operatorname{Cov}\{f_\theta(X)\}.
 \label{fclnl-supp:eq:normalization}
\end{equation}
\revgreen{The encoder covariance $\Sigma_\theta$ may be singular; the head must produce a full-rank output covariance satisfying $A\Sigma_\theta A^\top=I_d$. This matrix constraint specifies the joint variances and covariances of the output coordinates. Unit norms for individual observations and a bound on total variance impose different constraints.}

For any centered, normalized output \(z\), define
\[
 B_z=\operatorname{Cov}\{\mathbb E[z(X)\mid U]\},\qquad
 W_z=\mathbb E\{\operatorname{Cov}[z(X)\mid U]\}.
\]
Then \(B_z+W_z=I_d\). We retain the original, uncorrected
two-view population criterion, with fixed \(\lambda>0\):
\begin{equation}
 J_\lambda(z)=
 \operatorname{tr}\!\left[
 (W_z/2+\lambda I_d)^{-1}(B_z+W_z/2)
 \right].
 \label{fclnl-supp:eq:raw-objective}
\end{equation}
The within-source term in the numerator and both factors of
\(1/2\) are part of the analyzed objective.

\paragraph*{\revgreen{Connection to the empirical trace}}
Let \((U_i,X_i^{(1)},X_i^{(2)})\) be iid triples, where each
\((U_i,X_i^{(a)})\) has the joint law of \((U,X)\), and the
two views are conditionally independent given \(U_i\).
For a fixed \((\theta,A)\), write
\[
 M_i=\{z(X_i^{(1)})+z(X_i^{(2)})\}/2,\quad
 \Delta_i=\{z(X_i^{(1)})-z(X_i^{(2)})\}/2.
\]
With \(\bar M_N=N^{-1}\sum_iM_i\), the empirical scatters are
\[
 \widehat\Sigma_W=N^{-1}\sum_i\Delta_i\Delta_i^\top,\qquad
 \widehat\Sigma_B=N^{-1}\sum_i(M_i-\bar M_N)(M_i-\bar M_N)^\top.
\]
Conditional independence gives
\(\mathbb E(\Delta_i\Delta_i^\top)=W_z/2\) and
\(\operatorname{Cov}(M_i)=B_z+W_z/2\). Finite second moments, the strong law,
and continuity of the ridge inverse yield
\[
 \operatorname{tr}\{(\widehat\Sigma_W+\lambda I_d)^{-1}\widehat\Sigma_B\}
 \longrightarrow J_\lambda(z)\quad\text{almost surely}.
\]
This fixed-pair limit identifies the population objective. 

\subsection{A fixed signal space for joint nonlinear optimization}
\label{fclnl-supp:sec:function-space}

Let \(\mathcal V\) be a fixed closed subspace of the real Hilbert space
\(L^2_0(P_X)\), with inner product
\(\langle h,k\rangle=\mathbb E[h(X)k(X)]\).
\fclnladd{Require every centered output coordinate allowed throughout joint
optimization to belong to the same \(\mathcal V\). One may take
\(\mathcal V=L^2_0(P_X)\).}

Define the bounded operator
\begin{equation}
 T_{\mathcal V}:\mathcal V\longrightarrow L^2_0(P_U),\qquad
 T_{\mathcal V}h=\mathbb E[h(X)\mid U],\qquad \mathcal K_{\mathcal V}=T_{\mathcal V}^*T_{\mathcal V}.
 \label{fclnl-supp:eq:operator}
\end{equation}
Conditional Jensen's inequality gives \(\|T_{\mathcal V}h\|_2\le\|h\|_2\),
so \(\mathcal K_{\mathcal V}\) is a self-adjoint positive contraction. Its bilinear
form is
\begin{equation}
 \langle h,\mathcal K_{\mathcal V} k\rangle
 =\mathbb E\!\left[\mathbb E(h\mid U)\mathbb E(k\mid U)\right].
 \label{fclnl-supp:eq:operator-form}
\end{equation}
For the full space \(\mathcal V=L^2_0(P_X)\),
\(\mathcal K_{\mathcal V} h=\mathbb E\{\mathbb E[h(X)\mid U]\mid X\}\).
For a proper closed subspace, this expression must additionally
be projected onto \(\mathcal V\). This operator is fixed by the
source--view distribution and the chosen ambient function space,
even while \(\theta\) changes.

\paragraph*{\revgreen{Spectral separation assumption}}
Assume there are orthonormal functions
\(\phi_1,\ldots,\phi_d\in\mathcal V\) such that
\begin{equation}
 \mathcal K_{\mathcal V}\phi_i=\mu_i\phi_i,\qquad
 1\ge\mu_1\ge\cdots\ge\mu_d>\nu\ge0,
 \label{fclnl-supp:eq:spectral-block}
\end{equation}
and, for \(S_\star=\operatorname{span}\{\phi_1,\ldots,\phi_d\}\),
\begin{equation}
 \langle h,\mathcal K_{\mathcal V} h\rangle\le\nu\|h\|_2^2
 \quad\text{for all }h\in\mathcal V\cap S_\star^\perp.
 \qquad \gamma:=\mu_d-\nu>0.
 \label{fclnl-supp:eq:complement-bound}
\end{equation}
\revgreen{These assumptions specify a leading block and bound the quadratic form on its orthogonal complement. They apply to bounded positive contractions, including noncompact operators with the stated separated leading block.}

\fclnladd{Write \(\Phi(X)=(\phi_1(X),\ldots,\phi_d(X))^\top\), and define
the target class
\begin{equation}
 \mathcal H_\star
 =\bigcup_{q=1}^{\infty}
 \left\{C\Phi:C\in\mathbb R^{q\times d},\ \|C\|_{\mathrm{op}}\le1\right\}.
 \label{fclnl-supp:eq:targets}
\end{equation}
Here \(d\) is both the final-output dimension and the dimension of
the selected signal space, \(q\) is the finite number of target
functions, and each row of \(C\) gives one target's coefficients
in the basis \(\Phi\). The matrix \(C\) defines the targets and is
distinct from the projection head \(A\). The functions \(\phi_i\)
may be nonlinear in the raw input. The space \(S_\star\), and
hence \(\mathcal H_\star\), is fixed independently of the
representation being evaluated. The choice \(q=d\), \(C=I_d\)
includes recovery of the whole vector \(\Phi\); its recovery gives
every linear combination in \(S_\star\). The operator-norm
constraint fixes the scale of the target family for a uniform bound.
} \fclnladd{Alignment of the chosen scientific targets with this space is
a model condition.}

Let \(\mathfrak Z\) denote a nonempty class of outputs
\eqref{fclnl-supp:eq:architecture} satisfying \eqref{fclnl-supp:eq:normalization} and
the \(\mathcal V\)-membership condition. Exact spectral realizability
means that at least one pair \((\theta_\star,A_\star)\) produces
\(z=Q\Phi\) for an orthogonal \(d\times d\) matrix \(Q\).
This condition concerns the capacity of the joint architecture.
Subsection~\ref{fclnl-supp:sec:approximation} treats
approximate realizability.

\subsection{Joint encoder--head recovery theorem}
\label{fclnl-supp:sec:main-theorem}

\begin{proposition}[Spectral form of the original score]
\label{fclnl-supp:prop:score}
Let \(z_1,\ldots,z_d\) be centered orthonormal functions in \(\mathcal V\).
Then \((B_z)_{ij}=\langle z_i,\mathcal K_{\mathcal V} z_j\rangle\),
\(W_z=I_d-B_z\), and the eigenvalues
\(\theta_1(z)\ge\cdots\ge\theta_d(z)\) of \(B_z\) belong to
\([0,1]\). Moreover,
\begin{equation}
 J_\lambda(z)=\sum_{i=1}^d g_\lambda(\theta_i(z)),\qquad
 g_\lambda(t)=\frac{1+t}{1+2\lambda-t},
 \label{fclnl-supp:eq:spectral-score}
\end{equation}
where
\begin{equation}
 m_\lambda:=\frac{2(1+\lambda)}{(1+2\lambda)^2}
 \le g_\lambda'(t)
 =\frac{2(1+\lambda)}{(1+2\lambda-t)^2}
 \le M_\lambda:=\frac{1+\lambda}{2\lambda^2}.
 \label{fclnl-supp:eq:derivative}
\end{equation}
\end{proposition}

\begin{proof}
Equation~\eqref{fclnl-supp:eq:operator-form} gives the entries of \(B_z\).
The isometry \(b\mapsto\sum_jb_jz_j\) and
\(0\preceq\mathcal K_{\mathcal V}\preceq I\) give \(0\preceq B_z\preceq I_d\).
Total covariance gives \(W_z=I_d-B_z\). The two matrices in
\eqref{fclnl-supp:eq:raw-objective} are therefore
\(\{(1+2\lambda)I_d-B_z\}/2\) and \((I_d+B_z)/2\).
Diagonalizing \(B_z\) proves the score formula. Differentiation
and the endpoints \(t=0,1\) give the derivative bounds.
\end{proof}

For an output \(z\), let \(P_z\) be the orthogonal projection
in \(\mathcal V\) onto \(S_z=\operatorname{span}\{z_1,\ldots,z_d\}\).
Define
\begin{equation}
 L(z)=\sum_{i=1}^d\|(I-P_z)\phi_i\|_2^2,
 \qquad
 J_{\mathcal V}^\star=\sum_{i=1}^d g_\lambda(\mu_i),
 \qquad \delta_{\mathcal V}(z)=J_{\mathcal V}^\star-J_\lambda(z).
 \label{fclnl-supp:eq:ideal-deficit}
\end{equation}
The target readout errors from the head and backbone are
\begin{align}
 \mathcal E_{\boldsymbol h}(z)&=\inf_D\mathbb E\|\boldsymbol h-Dz\|_2^2,\notag\\
 \mathcal E_{\boldsymbol h}(f_\theta)&=\inf_B
   \mathbb E\|\boldsymbol h-B\widetilde f_\theta\|_2^2.
 \label{fclnl-supp:eq:errors}
\end{align}

\begin{theorem}[\fclnladd{Uniform recovery by a nonlinear encoder and linear head}]\color{black}
\label{fclnl-supp:thm:nonlinear}
Under \eqref{fclnl-supp:eq:spectral-block}--\eqref{fclnl-supp:eq:complement-bound},
for every admissible pair \((\theta,A)\),
\begin{equation}
 \delta_{\mathcal V}(z)\ge0,\qquad
 L(z)\le\frac{\delta_{\mathcal V}(z)}{m_\lambda\gamma},
 \label{fclnl-supp:eq:function-leakage}
\end{equation}
and uniformly over the target class \eqref{fclnl-supp:eq:targets},
\begin{align}
 \sup_{\boldsymbol h\in\mathcal H_\star}\mathcal E_{\boldsymbol h}(f_\theta)
 &\le\sup_{\boldsymbol h\in\mathcal H_\star}\mathcal E_{\boldsymbol h}(z)
 =L(z)\notag\\
 &\le\min\left\{d,
 \frac{\delta_{\mathcal V}(z)}{m_\lambda\gamma}\right\}.
 \label{fclnl-supp:eq:nonlinear-risk}
\end{align}
Over all orthonormal \(d\)-tuples in \(\mathcal V\), the maximum
score is \(J_{\mathcal V}^\star\), and equality holds exactly when
\(S_z=S_\star\).
If exact spectral realizability holds for \(\mathfrak Z\), the same
maximum is attained in the joint encoder--head class.
Every joint global maximizer then has \(z=Q\Phi\) for some
orthogonal \(Q\), and permits simultaneous exact linear recovery
of all targets in \(\mathcal H_\star\) from both its head output
and its encoder output.
\end{theorem}

\begin{proof}
We first show \(\theta_i(z)\le\mu_i\). The matrix \(B_z\)
is the compression of \(\mathcal K_{\mathcal V}\) to the \(d\)-dimensional space
\(S_z\). Its \(i\)th eigenvalue has the variational expression
\[
 \theta_i(z)=\max_{\substack{V\subset S_z\\\dim V=i}}
      \min_{\substack{h\in V\\\|h\|_2=1}}
      \langle h,\mathcal K_{\mathcal V} h\rangle.
\]
Every \(i\)-dimensional \(V\) contains a unit vector orthogonal
to \(\phi_1,\ldots,\phi_{i-1}\). Decompose that vector into
\(\operatorname{span}\{\phi_i,\ldots,\phi_d\}\) and
\(S_\star^\perp\). The cross terms vanish by self-adjointness
and the eigenfunction equations. Its Rayleigh quotient is at
most \(\mu_i\), because \(\nu<\mu_d\le\mu_i\).
This proves the compression inequality. The score formula and
\(g_\lambda'\ge m_\lambda\) now give
\begin{equation}
 \delta_{\mathcal V}(z)
 \ge m_\lambda\left\{\sum_{i=1}^d\mu_i-\operatorname{tr}(B_z)\right\}.
 \label{fclnl-supp:eq:score-trace-gap}
\end{equation}
The tuple \(\Phi\) attains \(J_{\mathcal V}^\star\).

Let \(P_\star\) project onto \(S_\star\), and set
\(a_i=\sum_{j=1}^d|\langle\phi_i,z_j\rangle|^2
=\|P_z\phi_i\|_2^2\). Bessel's inequality gives
\(0\le a_i\le1\), and
\[
 L(z)=d-\sum_i a_i
 =\sum_{j=1}^d\|(I-P_\star)z_j\|_2^2.
\]
Decomposing each \(z_j\) into its \(S_\star\) and orthogonal
components, and using \eqref{fclnl-supp:eq:complement-bound}, yields
\[
 \operatorname{tr}(B_z)=\sum_j\langle z_j,\mathcal K_{\mathcal V} z_j\rangle
 \le\sum_i\mu_i a_i+\nu L(z).
\]
Consequently,
\[
 \sum_i\mu_i-\operatorname{tr}(B_z)
 \ge\sum_i\mu_i(1-a_i)-\nu L(z)
 \ge(\mu_d-\nu)L(z).
\]
Combining this with \eqref{fclnl-supp:eq:score-trace-gap} proves
\eqref{fclnl-supp:eq:function-leakage}. \revgreen{The argument uses finite sums of quadratic forms of the bounded operator $\mathcal K_{\mathcal V}$.}

\fclnladd{For the readout, let \(R=\mathbb E(\Phi z^\top)\), and consider any
finite \(q\) and \(C\in\mathbb R^{q\times d}\), with
\(\boldsymbol h=C\Phi\).
Because \(\mathbb E(zz^\top)=I_d\), the unique best coefficient is
\(D_{\boldsymbol h}=\mathbb E(\boldsymbol h z^\top)=CR\). Completing the square gives
\[
 \mathbb E\|\boldsymbol h-Dz\|_2^2
 =\mathbb E\|\boldsymbol h-CRz\|_2^2+\|D-CR\|_{\mathrm{F}}^2.
\]
The residual covariance for \(\Phi\) is
\(M_z=I_d-RR^\top\), a positive semidefinite matrix with
\(0\preceq M_z\preceq I_d\) and trace \(L(z)\). Thus
\[
 \mathcal E_{C\Phi}(z)=\operatorname{tr}(CM_zC^\top).
\]
If \(\|C\|_{\mathrm{op}}\le1\), then \(C^\top C\preceq I_d\), so
\(\operatorname{tr}(CM_zC^\top)\le\operatorname{tr}(M_z)=L(z)\).
The choice \(q=d\), \(C=I_d\) attains equality. Consequently,
\[
 \sup_{\boldsymbol h\in\mathcal H_\star}\mathcal E_{\boldsymbol h}(z)
 =L(z)\le d.
\]
Every head readout \(Dz=DA\widetilde f_\theta\) is also a
backbone readout. Taking suprema and applying
\eqref{fclnl-supp:eq:function-leakage} proves
\eqref{fclnl-supp:eq:nonlinear-risk}. For an arbitrary coefficient
matrix, the same calculation also gives
\[
 \mathcal E_{C\Phi}(f_\theta)\le\mathcal E_{C\Phi}(z)
 \le\min\left\{\|C\|_{\mathrm F}^2,
 \|C\|_{\mathrm{op}}^2L(z)\right\}.
\]
The first bound uses \(M_z\preceq I_d\). For a scalar target
\(c^\top\Phi\) with \(\|c\|_2\le1\), the worst-case error is
\(\lambda_{\max}(M_z)\); the supremum in
\revgreen{\eqref{fclnl-supp:eq:nonlinear-risk} measures total error over vector targets.}}

If a tuple attains \(J_{\mathcal V}^\star\), the leakage is zero,
so every \(\phi_i\) belongs to \(S_z\). Both spaces have
dimension \(d\), hence \(S_z=S_\star\). Conversely, an
orthonormal basis of \(S_\star\) has compression eigenvalues
\(\mu_1,\ldots,\mu_d\) and attains the bound. Two orthonormal
bases of this space differ by an orthogonal matrix. Exact
realizability transfers this characterization to \(\mathfrak Z\).
\fclnladd{In that case \(\Phi=Q^\top z\) almost surely. Composing this
single vector identity with any \(C\) proves simultaneous exact
recovery throughout \(\mathcal H_\star\), from both outputs.}
\end{proof}

\subsection{Approximation error and near-optimal population loss}
\label{fclnl-supp:sec:approximation}

Write
\begin{equation}
 J_{\mathfrak Z}^\star=\sup_{z\in\mathfrak Z}J_\lambda(z),\qquad
 a_{\mathfrak Z}=J_{\mathcal V}^\star-J_{\mathfrak Z}^\star,\qquad
 \eta(z)=J_{\mathfrak Z}^\star-J_\lambda(z).
 \label{fclnl-supp:eq:class-deficits}
\end{equation}
\revgreen{Both deficits are nonnegative for every nonempty encoder--head class, and}
\(\delta_{\mathcal V}(z)=a_{\mathfrak Z}+\eta(z)\).
Theorem~\ref{fclnl-supp:thm:nonlinear} therefore gives
\fclnladd{\begin{align}
 \sup_{\boldsymbol h\in\mathcal H_\star}\mathcal E_{\boldsymbol h}(f_\theta)
 &\le\sup_{\boldsymbol h\in\mathcal H_\star}\mathcal E_{\boldsymbol h}(z)
 =L(z)\notag\\
 &\le\min\left\{d,
 \frac{a_{\mathfrak Z}+\eta(z)}{m_\lambda\gamma}\right\}.
 \label{fclnl-supp:eq:approximation-risk}
\end{align}}
\fclnladd{The next result bounds \(a_{\mathfrak Z}\) by how accurately the
joint architecture can approximate the leading eigenfunctions.}

\begin{corollary}[Approximate eigenfeatures and a linear whitening head]
\label{fclnl-supp:cor:approximation}
Suppose an encoder from the admissible class and a linear map
produce a centered vector
\(y=B\widetilde f_\theta\in\mathcal V^d\) such that
\begin{equation}
 \mathbb E\|y-\Phi\|_2^2\le\epsilon_{\rm app}^2,
 \qquad 0\le\epsilon_{\rm app}<1.
 \label{fclnl-supp:eq:feature-approximation}
\end{equation}
Suppose the class defining \(\mathfrak Z\) includes every linear head
for this encoder that satisfies the stated output constraints. Then
\begin{equation}
 a_{\mathfrak Z}\le M_\lambda\mu_1\epsilon_{\rm app}^2.
 \label{fclnl-supp:eq:approximation-deficit}
\end{equation}
In particular, arbitrarily accurate approximation of \(\Phi\)
by such centered outputs implies \(a_{\mathfrak Z}=0\).
\revgreen{Exact attainment by a finite architecture is covered by the spectral realizability condition.}
\end{corollary}

\begin{proof}
Set \(G=\mathbb E(yy^\top)\). For any \(b\in\mathbb R^d\),
\[
 \|b^\top y\|_2
 \ge\|b^\top\Phi\|_2-\|b^\top(y-\Phi)\|_2
 \ge(1-\epsilon_{\rm app})\|b\|_2.
\]
Thus \(G\succeq(1-\epsilon_{\rm app})^2I_d\), so the linear
head can form \(z=G^{-1/2}y\in\mathfrak Z\). Explicitly,
\(A_{\rm new}=G^{-1/2}B\) and \(G=B\Sigma_\theta B^\top\), so
\(A_{\rm new}\Sigma_\theta A_{\rm new}^\top=I_d\).
The new coordinates remain in \(\mathcal V\) by linearity. Since
\(S_z=\operatorname{span}\{y_1,\ldots,y_d\}\), projection
optimality gives
\(L(z)\le\sum_i\|\phi_i-y_i\|_2^2\le\epsilon_{\rm app}^2\).
As in the main proof, \(\theta_i(z)\le\mu_i\). The upper
slope bound in \eqref{fclnl-supp:eq:derivative} gives
\[
 \delta_{\mathcal V}(z)
 \le M_\lambda\{\textstyle\sum_i\mu_i-\operatorname{tr}(B_z)\}.
\]
Positivity of \(\mathcal K_{\mathcal V}\) on \(S_\star^\perp\) also gives
\(\operatorname{tr}(B_z)\ge\sum_i\mu_i a_i\). Hence the quantity in
braces is at most \(\mu_1L(z)\). Finally
\(a_{\mathfrak Z}\le\delta_{\mathcal V}(z)\), proving the claim.
\end{proof}

\begin{corollary}[Population negative-log-loss deficit]
\label{fclnl-supp:cor:log}
Let \(\ell_\lambda(z)=-\log J_\lambda(z)\), and suppose
\[
 \ell_\lambda(z)-\inf_{v\in\mathfrak Z}\ell_\lambda(v)\le\epsilon.
\]
Then
\begin{equation}
 \eta(z)\le J_{\mathfrak Z}^\star(1-e^{-\epsilon})
 \le\frac d\lambda(1-e^{-\epsilon})
 \le\frac d\lambda\epsilon.
 \label{fclnl-supp:eq:log-deficit}
\end{equation}
\fclnladd{Consequently, the uniform readout bound holds with
\(a_{\mathfrak Z}+d\epsilon/\lambda\) in place of \(\delta_{\mathcal V}(z)\).}
\end{corollary}

\begin{proof}
All scores are positive because \(g_\lambda(t)\ge g_\lambda(0)>0\).
The loss condition implies
\(J_\lambda(z)\ge J_{\mathfrak Z}^\star e^{-\epsilon}\).
Also \(g_\lambda(t)\le g_\lambda(1)=1/\lambda\), and
\(1-e^{-\epsilon}\le\epsilon\).
\end{proof}

\subsection{More output dimensions than signal directions}
\label{fclnl-supp:sec:rank}

The next result replaces the leading-block assumptions
\eqref{fclnl-supp:eq:spectral-block}--\eqref{fclnl-supp:eq:complement-bound} by
a finite-rank condition. \fclnladd{For \(d>r\), recovery is controlled by the smallest positive eigenvalue.}

\begin{corollary}[\fclnladd{Finite-rank nonlinear signal operator}]\color{black}
\label{fclnl-supp:cor:rank}
Suppose \(\mathcal K_{\mathcal V}\) has rank \(r\ge1\), with positive eigenvalues
\(\mu_1\ge\cdots\ge\mu_r>0\) and orthonormal eigenfunctions
\(\phi_1,\ldots,\phi_r\). Let \(r\le d\le\dim\mathcal V\).
Assume the joint class can realize an orthonormal \(d\)-tuple
whose span contains \(S_r=\operatorname{range}(\mathcal K_{\mathcal V})\). Then
\begin{equation}
 J_{\mathfrak Z}^\star=\sum_{i=1}^r g_\lambda(\mu_i)
                  +(d-r)g_\lambda(0).
 \label{fclnl-supp:eq:rank-optimum}
\end{equation}
Every global maximizer has \(S_r\subseteq S_z\), and every
admissible output with this inclusion is a global maximizer.
Write \(\Phi_r=(\phi_1,\ldots,\phi_r)^\top\), and define
\[
 \mathcal H_r=\bigcup_{q=1}^{\infty}
 \left\{C\Phi_r:C\in\mathbb R^{q\times r},\ \|C\|_{\mathrm{op}}\le1\right\}.
\]
Let \(L_r(z)=\sum_{i=1}^r\|(I-P_z)\phi_i\|_2^2\).
For every admissible pair,
\begin{align}
 \sup_{\boldsymbol h\in\mathcal H_r}\mathcal E_{\boldsymbol h}(f_\theta)
 &\le\sup_{\boldsymbol h\in\mathcal H_r}\mathcal E_{\boldsymbol h}(z)
 =L_r(z)\notag\\
 &\le\min\left\{r,
 \frac{J_{\mathfrak Z}^\star-J_\lambda(z)}
      {m_\lambda\mu_r}\right\}.
 \label{fclnl-supp:eq:rank-risk}
\end{align}
Thus every global maximizer permits simultaneous exact linear
recovery of all targets in \(\mathcal H_r\) from both outputs.
\end{corollary}

\begin{proof}
Compression gives \(\theta_i(z)\le\mu_i\) for \(i\le r\)
and \(\theta_i(z)=0\) for \(i>r\). Completing an orthonormal
basis of \(S_r\) by \(d-r\) functions in \(\ker\mathcal K_{\mathcal V}\) attains
\eqref{fclnl-supp:eq:rank-optimum}; realizability makes this attainable
in the joint class. With \(a_i=\|P_z\phi_i\|_2^2\),
\[
 J_{\mathfrak Z}^\star-J_\lambda(z)
 \ge m_\lambda\sum_{i=1}^r\mu_i(1-a_i)
 \ge m_\lambda\mu_r\sum_{i=1}^r(1-a_i).
\]
\fclnladd{Zero deficit is equivalent to all the \(\phi_i\) belonging
to \(S_z\). With the rectangular matrix
\(R=\mathbb E(\Phi_r z^\top)\in\mathbb R^{r\times d}\), the readout
calculation in the main proof gives
\(\mathcal E_{C\Phi_r}(z)=\operatorname{tr}\{C(I_r-RR^\top)C^\top\}\).
The residual covariance satisfies \(0\preceq I_r-RR^\top\preceq I_r\)
and has trace \(L_r(z)\). Since \(C^\top C\preceq I_r\), the
supremum equals \(L_r(z)\), attained by \(q=r\), \(C=I_r\).
This gives the cap \(r\); the score-deficit inequality above and
composition with the linear head prove \eqref{fclnl-supp:eq:rank-risk}.
At zero deficit, \(\Phi_r=Rz\) almost surely, so one vector
identity yields exact recovery for every target in \(\mathcal H_r\).}
\end{proof}

When \(d>r\), the additional functions may lie anywhere in
\(\ker\mathcal K_{\mathcal V}\); \fclnladd{the optimal output space is specified up to the choice of these kernel directions.}
When \(r=0\), every normalized tuple has score
\(d g_\lambda(0)\), \fclnladd{and the positive signal space is empty.}
If \(d<r\), the canonical signal target \(\Phi_r\) has
readout error at least \(r-d\), by Bessel's inequality.
If \(\mathcal V\) is finite-dimensional and \(d=\dim\mathcal V\), every
normalized output basis spans \(\mathcal V\).
\fclnladd{The rank of the nonlinear operator \(\mathcal K_{\mathcal V}\) depends on
the conditional-mean function space, even for finite-dimensional
latent variables.}

\subsection{\fclnladd{A nonlinear example with a spectral gap}}
\label{fclnl-supp:sec:example}

Let \(U_1,U_2\) be independent equiprobable signs. For each
view draw independent signs \(\xi_1,\xi_2\), independent of
\(U\), with
\[
 \Pr(\xi_j=1)=(1+\rho_j)/2,\qquad
 0<\rho_2\le\rho_1<1.
\]
Set
\begin{equation}
 S_j=U_j\xi_j,\qquad X=(S_1,S_1S_2).
 \label{fclnl-supp:eq:nonlinear-example}
\end{equation}
Fresh independent flips conditional on \(U=(U_1,U_2)\)
generate the paired augmented views.

The coordinates \(S_1,S_2\) are independent equiprobable signs.
On the four-point input space, \(\mathcal V=L^2_0(P_X)\) has the
orthonormal basis
\[
 \phi_1(X)=S_1=X_1,\qquad
 \phi_2(X)=S_2=X_1X_2,\qquad
 \phi_3(X)=S_1S_2=X_2.
\]
Their conditional expectations are
\(\rho_1U_1\), \(\rho_2U_2\), and
\(\rho_1\rho_2U_1U_2\), respectively. These are orthogonal
in \(L^2(P_U)\). Equation~\eqref{fclnl-supp:eq:operator-form} therefore
gives the eigenvalues
\[
 \mu_1=\rho_1^2,\quad \mu_2=\rho_2^2,\quad
 \mu_3=\rho_1^2\rho_2^2.
\]
For \(d=2\), the gap is
\(\gamma=\rho_2^2(1-\rho_1^2)>0\).
A realizable nonlinear encoder and linear head are
\[
 f_{\theta_\star}(x)=(x_1,x_1x_2),\qquad A_\star=I_2.
\]
The resulting output \((S_1,S_2)\) has identity covariance
and attains \(J_{\mathcal V}^\star=g_\lambda(\rho_1^2)+g_\lambda(\rho_2^2)\).
\fclnladd{\fclnladd{The theorem implies that every joint maximizer recovers
the entire selected space \(\operatorname{span}\{S_1,S_2\}\) by
linear readout, with a uniform guarantee over \(\mathcal H_\star\).}}

For comparison, the feasible identity encoder \(f(x)=(x_1,x_2)\)
with \(A=I_2\) yields \((S_1,S_1S_2)\). Its optimal linear
prediction error for \(S_2\) is one, and its score deficit is
\[
 g_\lambda(\rho_2^2)-g_\lambda(\rho_1^2\rho_2^2)>0.
\]
The missing target is a nonlinear function of the raw input,
and the spectral objective distinguishes the two encoders.
\fclnladd{\fclnladd{The full conditional-mean operator has rank three; the
leading-block theorem recovers the selected two-dimensional
signal space in the two-dimensional output.}}

\paragraph*{\revgreen{Compatibility with central targets}}
The input determines \(S\), and
\(\mathbb E(U\mid X)=(\rho_1S_1,\rho_2S_2)^\top\).
Any sufficient sub-sigma-field \(\mathcal G\subseteq\sigma(X)\)
therefore makes \(S\) measurable, since both \(\rho_j>0\).
Thus the smallest sufficient sigma-field is \(\sigma(S)=\sigma(X)\).
The targets \(S_j\) are centered, source-informative members
of the central function class $\mathcal C$ in Assumption~\ref{ass:central-sigma}.

More generally, if \(\mathcal V=L^2_0(P_X)\) and a central sufficient
sigma-field \(\mathcal G\) exists, every positive-eigenvalue
function of \(\mathcal K_{\mathcal V}\) belongs to \(L^2_0(\mathcal G)\).
Indeed, sufficiency makes \(\mathbb E[q(U)\mid X]\)
\(\mathcal G\)-measurable for any square-integrable \(q\),
and \(\phi_i=\mu_i^{-1}T_{\mathcal V}^*T_{\mathcal V}\phi_i\).
Also \(\|T_{\mathcal V}\phi_i\|_2^2=\mu_i>0\), so these eigenfunctions
are source-informative. \fclnladd{For a proper ambient subspace,
central-target membership depends on the extra projection in
\(T_{\mathcal V}^*\) and must be verified for the chosen space.}

\subsection{\fclnladd{Recovery guarantees and their conditions}}
\label{fclnl-supp:sec:scope}

\paragraph*{\revgreen{Recovery conclusion}}
\fclnladd{Under population covariance normalization, spectral separation,
and realizability of the selected signal eigenspace, every
population-optimal nonlinear encoder--linear head pair permits
simultaneous exact linear recovery throughout \(\mathcal H_\star\).
Near-optimal pairs satisfy a uniform linear-readout error bound determined
by their population objective deficit, approximation error, and
the spectral gap. Because the head is linear, the same upper
bound applies to readout from the encoder output. Recovery of the
whole selected space means recovery of every linear combination
of its eigenfunctions; the error bound uses the normalization
\(\|C\|_{\mathrm{op}}\le1\).}

For a unit-variance scalar target in \(S_\star\), a head with
readout error at least \(\tau>0\) must satisfy
\(\delta_{\mathcal V}(z)\ge m_\lambda\gamma\tau\).
Thus exact global optimality excludes linear suppression of
these targets under realizability, while near-optimality
controls its magnitude. Positive but small error still counts
as suppression under a definition that uses a zero threshold.

\paragraph*{\revgreen{Spectral separation}}
Take independent equiprobable signs \(S_1,S_2\), noiseless
views \(U=X=(S_1,S_2)\), and \(d=2\).
Both encoders \((S_1,S_2)\) and \((S_1,S_1S_2)\), with
identity heads, have covariance \(I_2\) and maximum score
\(2/\lambda\). The latter has linear-readout MSE one for
\(S_2\). Here the full \(\mathcal K_{\mathcal V}\) is the identity on the
three-dimensional centered function space, so the required
boundary gap is absent. 

\paragraph*{\revgreen{Target alignment}}
\fclnladd{\fclnladd{The recovery guarantee is uniform over the normalized target
class \(\mathcal H_\star\) in the selected leading function space.
Applying it to a scientific target requires that target to belong
to this space. Nonlinear functions of one strong factor
can compete with a weaker factor. Assessing target coverage
therefore requires the spectral dimension and separation, even
when the number of latent factors is known. Other nonlinear
functions of the retained signals require their own linear-readout
recovery analysis.}}

\paragraph*{\revgreen{Covariance normalization}}
\fclnladd{A total-variance bound admits duplicated output coordinates.}
For a whitened feature vector \(F\) with
\(\operatorname{Cov}\{\mathbb E(F\mid U)\}=\operatorname{diag}(1,\kappa,0)\),
\(0<\kappa<1\), the two-row head that duplicates \(F_1\)
has total variance two, zero within-source covariance, and
score \(2/\lambda\). It cannot recover \(F_2\).
It is globally optimal among all outputs of total variance
at most two, since
\[
 J_\lambda(z)\le\lambda^{-1}\operatorname{tr}(B_z+W_z/2)
 \le\lambda^{-1}\operatorname{tr}\{\operatorname{Cov}(z)\}\le2/\lambda.
\]
\revgreen{The duplicated output has singular covariance and lies outside the feasible set $\operatorname{Cov}(z)=I_2$. For a fixed isotropic ridge, covariance normalization changes the optimization problem. Sample whitening and batch-dependent layers introduce additional statistical dependence.}

\paragraph*{\revgreen{Population optimization and empirical training}}
\revgreen{Applying this population theorem to an SGD-trained network requires bounds on its population deficit and approximation error, together with the stated population covariance normalization. The fixed-pair limit in Subsection~\ref{fclnl-supp:sec:architecture} identifies the population score. A fitted-pair analysis also accounts for optimization and data-dependent selection. The uniform recovery statement ranges over targets in a fixed population signal space. The backbone can retain features discarded by the linear head, allowing smaller readout error at the backbone.}

\paragraph*{\revgreen{Relation to the preceding theory}}
\fclnladd{\fclnladd{The proof derives uniform recovery bounds over a selected signal space
directly from the original population FCL score under normalization
and separation, with exact recovery under realizability. The spectral argument is independent
of central-projection monotonicity and SDR completeness. The
margin-selected eigenvalue objective in Section~\ref{app:imple}
and losses with additional penalties require a separate analysis.}}

} % End of the red nonlinear recovery addition.

% END INCORPORATED SUPPLEMENT BODY
\clearpage
\bibliographystyle{imsart-nameyear}
\bibliography{ref}
\end{document}